%% file: arxiv.tex
\documentclass[10pt,twocolumn,letterpaper]{article}

\usepackage[pagenumbers]{cvpr} 

\input{math_commands.tex}
\usepackage{graphicx}
\usepackage{amsmath}
\usepackage{amssymb}
\usepackage{amsfonts}
\usepackage{bm}
\usepackage{booktabs}
\usepackage[T1]{fontenc}
\usepackage{color}
\usepackage{multirow}
\usepackage{multicol}
\usepackage{tipa}
\usepackage{bbding}
\usepackage{pifont}
\usepackage{mathptmx} 
\usepackage{caption}
\usepackage{subcaption}

\input{math_commands.tex}
\usepackage{algorithm} 
\usepackage{algorithmic}

\usepackage[algo2e,ruled,vlined]{algorithm2e}

\renewcommand\footnotemark{}

\usepackage[pagebackref,breaklinks,colorlinks]{hyperref}

\usepackage[capitalize]{cleveref}
\crefname{section}{Sec.}{Secs.}
\Crefname{section}{Section}{Sections}
\Crefname{table}{Table}{Tables}
\crefname{table}{Tab.}{Tabs.}

\def\cvprPaperID{1092} 
\def\confName{CVPR}
\def\confYear{2023}

\begin{document}

\title{Memories are One-to-Many Mapping Alleviators in Talking Face Generation}

\author{
	Anni Tang$^1$\thanks{Corresponding authors: Xu Tan and Li Song (song\_li@sjtu.edu.cn). Project page: \href{https://memoryface.github.io}{memoryface.github.io}.}, Tianyu He, Xu Tan, Jun Ling$^1$, and Li Song$^1$\\
	$^1$Shanghai Jiao Tong University \\
{\tt\small \{memory97,lingjun,song\_li\}@sjtu.edu.cn}
}
\maketitle

\vspace{-2mm}

\begin{abstract}
    \vspace{-1mm}
    Talking face generation aims at generating photo-realistic video portraits of a target person driven by input audio. Due to its nature of one-to-many mapping from the input audio to the output video (\eg, one speech content may have multiple feasible visual appearances), learning a deterministic mapping like previous works brings ambiguity during training, and thus causes inferior visual results. Although this one-to-many mapping could be alleviated in part by a two-stage framework (\ie, an audio-to-expression model followed by a neural-rendering model), it is still insufficient since the prediction is produced without enough information (\eg, emotions, wrinkles, \etc.). In this paper, we propose MemFace to complement the missing information with an implicit memory and an explicit memory that follow the sense of the two stages respectively. More specifically, the implicit memory is employed in the audio-to-expression model to capture high-level semantics in the audio-expression shared space, while the explicit memory is employed in the neural-rendering model to help synthesize pixel-level details. Our experimental results show that our proposed MemFace surpasses all the state-of-the-art results across multiple scenarios consistently and significantly.
\end{abstract}


\vspace{-3mm}
\section{Introduction}
\label{sec:intro}
Talking face generation enables synthesizing photo-realistic video portraits of a target person in line with the speech content~\cite{suwajanakorn2017synthesizing,chen2019hierarchical,thies2020neural,ji2021audio,guo2021ad,chen2020duallip,ling2022stableface,tang2022real}. It shows great potential in applications like virtual avatars, online conferences, and animated movies since it conveys the visual content of the interested person besides the audio.

\begin{figure}[t]
  \centering
  \includegraphics[width=0.95\linewidth]{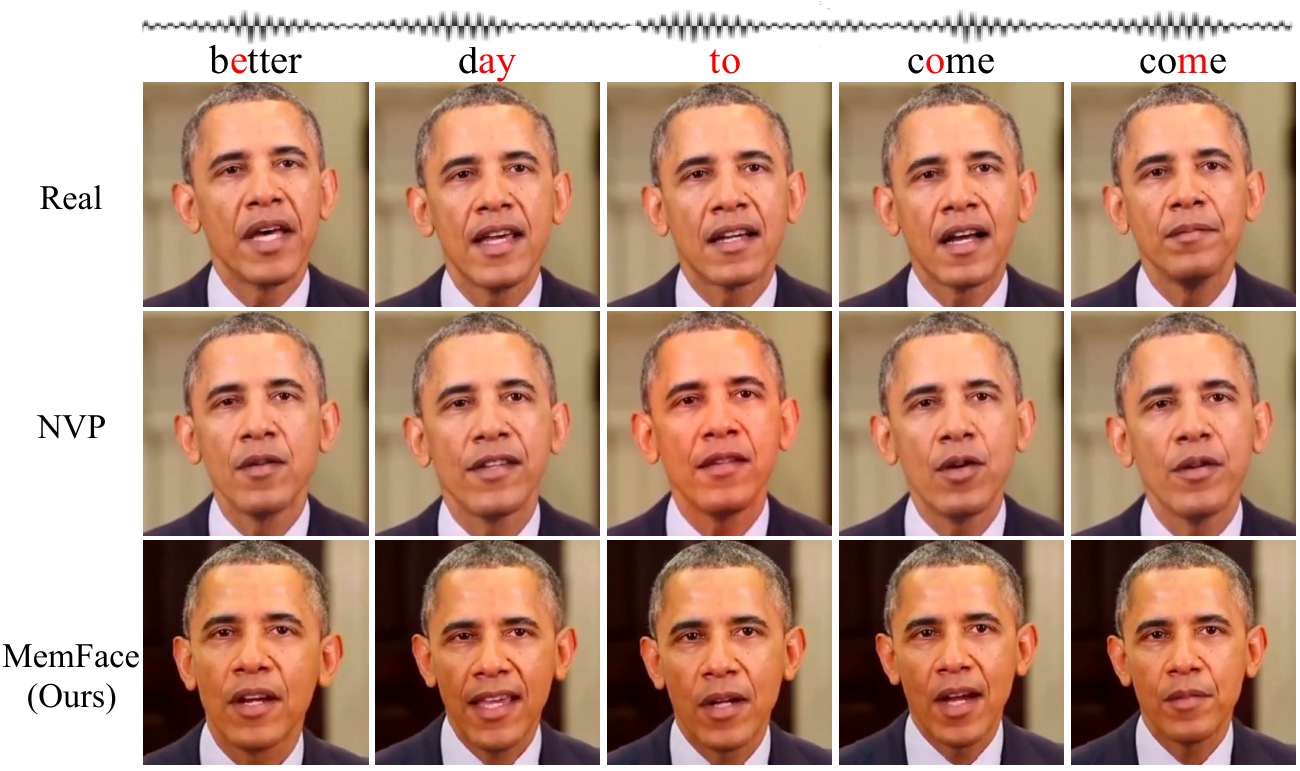}
  \vspace{-1mm}
  \caption{Example frames of real video (1st row), NVP~\cite{thies2020neural} (2nd row) and the proposed MemFace (3rd row) for saying `better day to come'.
  Compared to the baseline NVP~\cite{thies2020neural}, our results demonstrate higher lip-sync quality and more realistic rendering results by alleviating the one-to-many mapping problem with memories.}
  \vspace{-5mm}
  \label{fig:cover_single}
\end{figure}

The most popular methods to solve audio-driven talking face generation follow a two-stage framework~\cite{thies2020neural}, where an intermediate representation (\eg, 2D landmarks~\cite{suwajanakorn2017synthesizing,siarohin2019first,chen2019hierarchical,lu2021live}, blendshape coefficients of 3D face models~\cite{kim2018deep,thies2020neural,wen2020photorealistic,wu2021imitating}, \etc.) is first predicted from the input audio, then a renderer is employed to synthesize the video portraits according to the predicted representation. Along this path, remarkable progress has been made towards improving the overall realness of the video portraits, by achieving natural head movements~\cite{chen2020talking,yi2020audio,zhang2021facial,zhou2021pose}, enhancing lip-sync quality~\cite{prajwal2020lip,lahiri2021lipsync3d,park2022synctalkface}, generating emotional expression~\cite{ji2021audio,wu2021imitating,liang2022expressive}, \etc. However, the aforementioned methods are biased towards learning a deterministic mapping from the given audio to a video, while it is worth noting that talking face generation is inherently a one-to-many mapping problem. It means that, for an input audio clip, there are multiple feasible visual appearances of the target person due to the variations of phoneme contexts~\cite{agarwal2020detecting}, emotions~\cite{cudeiro2019capture}, and illumination conditions~\cite{smith2020morphable}, \etc. In this way, learning a deterministic mapping brings ambiguity during the training, making it harder to yield realistic visual results.

To some extent, this one-to-many mapping could be alleviated in part by the two-stage framework~\cite{thies2020neural,wen2020photorealistic,wu2021imitating}, since it decomposes the whole one-to-many mapping difficulty into two sub-problems (\ie, an audio-to-expression problem and a neural-rendering problem). However, although effective, each of these two stages is still optimized to predict the information that is missed by the input, thus remaining hard for prediction. For example, the audio-to-expression model learns to produce an expression that semantically matches the input audio, which misses the high-level semantics like habits, attitudes, \etc. While the neural-rendering model synthesizes the visual appearances based on the expression estimation, which misses the pixel-level details like wrinkles, shadows, \etc. To further alleviate the one-to-many mapping problem, in this paper, we propose MemFace to complement the missing information with memories~\cite{weston2014memory,sukhbaatar2015end}, by devising an implicit memory and an explicit memory that follows the sense of the two stages respectively. More specifically, the implicit memory is jointly optimized with the audio-to-expression model to complement the semantically-aligned information, while the explicit memory is constructed in a non-parametric way and tailored for each target person to complement visual details.

Therefore, instead of directly using the input audio to predict the expression~\cite{thies2020neural}, our audio-to-expression model leverages the extracted audio feature as the query to attend to the implicit memory. The attention result, which served as semantically-aligned information, is then complemented with the audio feature to yield expression output. By enabling end-to-end training, the implicit memory is encouraged to relate high-level semantics in the audio-expression shared space, thus narrowing the semantic gap between the input audio and the output expression.

After obtaining the expression, the neural-rendering model is employed to synthesize the visual appearances based on the mouth shapes that are obtained from expression estimations. To further complement pixel-level details between them, we first construct the explicit memory for each person by regarding the vertices of 3D face models~\cite{blanz1999morphable,wood2021fake} and its associated image patches as keys and values respectively. Then, for each input expression, its corresponding vertices are used as the query to retrieve the similar keys in the explicit memory, and the associated image patch is returned as the pixel-level details to the neural-rendering model. Intuitively, by introducing the explicit memory to the model, it allows the model to selectively associate expression-required details without generating them by the model itself, and thus eases the generation process.

Extensive experiments on various widely adopted datasets (\eg, Obama~\cite{suwajanakorn2017synthesizing}, HDTF~\cite{zhang2021flow}) verify that the proposed MemFace achieves state-of-the-art lip-sync and rendering quality, surpassing all the baseline methods across multiple scenarios consistently and significantly.
For example, our MemFace achieves a relative improvement of $37.52\%$ in the subjective evaluation of the Obama dataset.

\begin{figure*}[t]
  \centering
  \includegraphics[width=0.95\linewidth]{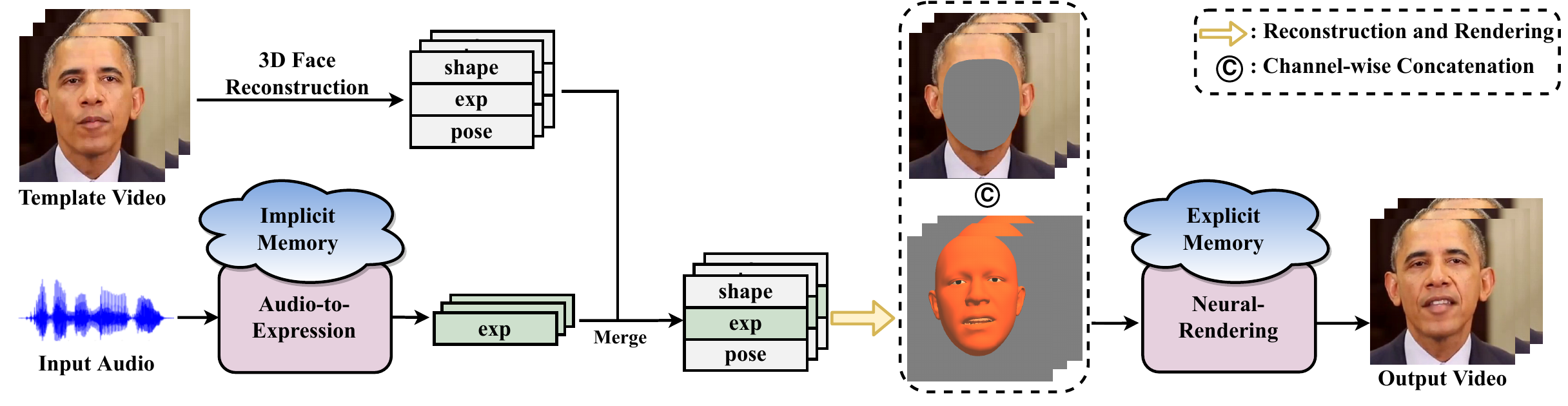}
  \vspace{-3mm}
  \caption{The overview of MemFace. To alleviate the one-to-many mapping difficulty, we propose to complement the missing information with memories. The implicit memory is introduced to the audio-to-expression model to complement the semantically-aligned information (see Fig.~\ref{fig:a2e_model}), while the explicit memory is introduced to the neural-rendering model to retrieve the personalized visual details (see Fig.~\ref{fig:rendering_model}).}
  \vspace{-3mm}
  \label{fig:pipeline}
\end{figure*}

\vspace{-1mm}
\section{Related Works}
\label{sec:related_work}
\vspace{-1mm}
\subsection{Talking Face Generation}
\vspace{-1mm}
Audio-driven talking face generation enables synthesizing photo-realistic video portraits in sync with the input speech content.
To tackle this problem, lots of methods have been proposed for improving the overall realness of the generated video portraits~\cite{thies2020neural,lu2021live,han2022show,meshry2021learned,hong2022depth}, such as to achieve natural head movements~\cite{chen2020talking,yi2020audio,zhang2021facial,zhou2021pose}, to enhance lip-sync~\cite{lahiri2021lipsync3d,park2022synctalkface,suwajanakorn2017synthesizing,thies2020neural,chen2019hierarchical,chen2022transformer}, to make emotional expression~\cite{ji2021audio,wu2021imitating,liang2022expressive}, \etc.
For example, Thies \etal~\cite{thies2020neural} proposed a two-stage framework that estimates lip motions and then renders the appearance of the target person.
Inspired by dynamic NeRF~\cite{mildenhall2021nerf,gafni2021dynamic}, recent advances also proposed methods that directly map the audio features to dynamic facial radiance fields for portrait rendering~\cite{guo2021ad,shen2022learning,liu2022semantic}.

Although these methods could alleviate the one-to-many mapping to some extent by the two-stage framework~\cite{kim2018deep,thies2020neural,wen2020photorealistic,wu2021imitating,suwajanakorn2017synthesizing,siarohin2019first,chen2019hierarchical,lu2021live} or the implicit function~\cite{guo2021ad,shen2022learning,liu2022semantic}, they were still optimized to predict the information that is missed by the input. Differently, we propose to complement the missing information with memories to tackle the one-to-many mapping difficulty.

\subsection{Memory-based Networks}

\vspace{-1mm}
We introduce the works that are related to implicit memory and explicit memory respectively in this section.

\vspace{-4mm}
\paragraph{Implicit memory.}
There were a number of attempts to introduce a specialized implicit memory that can be read and written for better memorization~\cite{weston2014memory,sukhbaatar2015end,kumar2016ask,lee2018memory}. They typically used continuous memory representations~\cite{sukhbaatar2015end} or key-value pairs~\cite{miller2016key} to read/write memories, allowing them to train the memory in an end-to-end way. Based on the success of implicit memory and the observation of the one-to-many mapping nature of audio-to-expression learning, we propose to incorporate an implicit memory into audio-to-expression stage to complement the missing semantically-aligned information and tackle the one-to-many mapping.

\vspace{-4mm}
\paragraph{Explicit memory.}
Augmenting neural networks with an explicit external memory has recently drawn attention in natural language processing~\cite{khandelwal2020generalization,khandelwal2021nearest,sheynin2022knn}.
For retrieval-based visual models, different from early attempts that only exploited the training data itself~\cite{long2022retrieval,siddiqui2021retrievalfuse,tseng2020retrievegan}, recent advances introduced an external memory for text-to-image generation~\cite{blattmann2022retrieval}. Unlike these approaches that built a unified memory for every sample, we build the explicit memory for each identity to complement the personalized visual details for realistic talking face generation.

It should be noted that Yi \etal~\cite{yi2020audio} and Park \etal~\cite{park2022synctalkface} also introduced memory to the talking face generation. Our solution is different from theirs in that: 1) Memory network in MemGAN~\cite{yi2020audio} stored paired spatial features (extracted from a pre-trained ResNet-18~\cite{he2016deep}) and identity features (extracted from a pre-trained ArcFace~\cite{deng2019arcface}) that provide more identity information to refine the rendering process. In contrast, we directly retrieve the mouth-related image patches from the video of the target person with our explicit memory, which makes the memory construction easier and complements the model with more visual details. 2) SyncTalkFace~\cite{park2022synctalkface} leveraged an audio-lip memory, where the audio features and lip features are regarded as keys and values respectively. Due to the large gap between audio and visual appearance, they used explicit constraints to align the two modalities. However, simultaneously aligning both the semantics and visual details in one memory is still challenging. In contrast, our MemFace employs an implicit memory and an explicit memory to complement the semantically-aligned information and pixel-level details respectively, therefore making the prediction easier.

\vspace{-1mm}
\section{MemFace}
\label{sec:method}

\vspace{-1mm}
\paragraph{Preliminaries.}

Our training corpus contains video portraits that are synced with the audio stream. For data preprocessing, following previous methods~\cite{cudeiro2019capture,sun2016phonetic,thies2020neural}, we transform the audio content to audio feature $\rmA$ with a pre-trained speech-to-text model~\cite{amodei2016deep}. The audio feature $\rmA$ is used as input instead of the raw audio waveform.

We adopt blendshape coefficients of 3D face model~\cite{blanz1999morphable,wood2021fake} as the intermediate representation due to its flexibility in talking face generation~\cite{kim2018deep,thies2020neural,wen2020photorealistic,wu2021imitating}. The 3D face model is a statistical model that enables the 3D reconstruction from the disentangled facial shape, expression, and other properties with delta-blendshapes~\cite{blanz1999morphable,wood2021fake}. We obtain the labels of shape coefficients $\boldsymbol{\alpha}_{id}$, expression coefficients $\boldsymbol{\alpha}_{exp}$, and pose coefficients $\boldsymbol{\alpha}_{pose}$ following Wood \etal~\cite{wood2021fake} in our experiments. To easily leverage the 3D face in our model, we project the 3D face onto the 2D image plane through a perspective camera model, resulting in an image $\rmI_{3D}$.

To fully exploit the coefficients of 3D face model, we also extract the coordinates of all vertices $\rmO$ by reconstructing the 3D face from the coefficients~\cite{wood2021fake}. Since the mouth-related region is the focus for talking, we only use the pre-defined mouth-related vertices $\rmO_{m}$. The obtained $\rmO_{m}$ is leveraged in both the audio-to-expression model and the neural-rendering model.

\vspace{-3mm}
\paragraph{Method overview.}

As shown in Fig.~\ref{fig:pipeline}, given input audio of the target person, our goal is to synthesize a photo-realistic video portrait that is consistent with the speech content. To achieve this goal, our inputs consist of an input audio feature $\rmA$ and a template video of the target person. For the template video, we follow the previous works~\cite{thies2020neural,wen2020photorealistic,yi2020audio} to mask the face region. To make the prediction focus on the face region, the remaining part of the template video is employed as input to produce extra information (see Fig.~\ref{fig:pipeline} for illustration).
We first employ our audio-to-expression model $f$ to take in the extracted audio feature $\rmA$, and predict the mouth-related expression coefficients $\boldsymbol{\hat{\alpha}}_{exp}$. The predicted expression coefficients $\boldsymbol{\hat{\alpha}}_{exp}$ are then merged with the original shape and pose coefficients of the template video, and yield an image $\hat{\rmI}_{3D}$ that corresponds to the predicted expression coefficients. Next, our neural-rendering model $g$ takes in the $\hat{\rmI}_{3D}$ and the masked template video, and outputs the final results that correspond to the mouth shape of $\hat{\rmI}_{3D}$. In this way, the audio-to-expression model is responsible for lip-sync quality, while the neural-rendering model is responsible for rendering quality.

However, this two-stage framework is still insufficient for tackling one-to-many mapping difficulty, since each stage is optimized to predict the information that is missed by the input (\eg, habits, wrinkles, \etc.). Therefore, we propose to complement the missing information with memory (see Sec.~\ref{sec:memory} for details). Furthermore, given that the audio-to-expression model $f$ and the neural-rendering model $g$ play different roles in talking face generation, we devise two variants of memory for complementing missing information accordingly. The details for applying memories to $f$ and $g$ are elaborated in Sec.~\ref{sec:a2e} and Sec.~\ref{sec:nr} respectively.

\vspace{-1mm}
\subsection{Memory as One-to-Many Mapping Alleviator}
\label{sec:memory}

\vspace{-1mm}
Memory enables an easy way to read and write to scalable storage for neural networks~\cite{weston2014memory,sukhbaatar2015end}. In this work, we propose to complement the missing information with memories, therefore alleviating the one-to-many mapping challenge. Formally, suppose that we have a key set $\rmK$ and a value set $\rmV$ that are to be stored in the memory, where each item of $\rmK$ is associated with a value in $\rmV$. For a query $\rmQ$ that is derived from the input, we compute the matching score between $\rmQ$ and key set $\rmK$ by taking a similarity function $\texttt{sim} (\cdot, \cdot)$, then relate different values in $\rmV$ via the similarity between $\rmQ$ and $\rmK$:
\begin{equation}
    \label{equ:mem}
    \texttt{attn} (\rmQ, \rmK, \rmV) = [\texttt{sim} (\rmQ, \rmK) \rmV \rmW_V] \rmW_O,
    \vspace{-0.1cm}
\end{equation}
where $\rmW_V$ and $\rmW_O$ denote the model parameters related to the memory mechanism.
A simple way to match the query $\rmQ$ and the key set $\rmK$ is to take the inner product followed by a softmax function~\cite{vaswani2017attention}:
\begin{equation}
    \label{equ:imp_mem}
    \texttt{sim} (\rmQ, \rmK) = \texttt{softmax} (\frac{\rmQ \rmW_Q (\rmK \rmW_K)^{\textbf{T}}}{\sqrt{h}}),
    \vspace{-0.1cm}
\end{equation}
where $\rmW_Q$ and $\rmW_K$ denote the model parameters, $h$ is the dimension of the hidden states.

In order to effectively complement the missing information in various scenarios, we design two memory variants that differ in the forms of the key set $\rmK$ and the value set $\rmV$. In Sec.~\ref{sec:discussion}, we also give insights into different variants.

\subsection{Complementing Expression Prediction with Implicit Memory}
\label{sec:a2e}

\begin{figure}[t]
  \centering
  \includegraphics[width=1\linewidth]{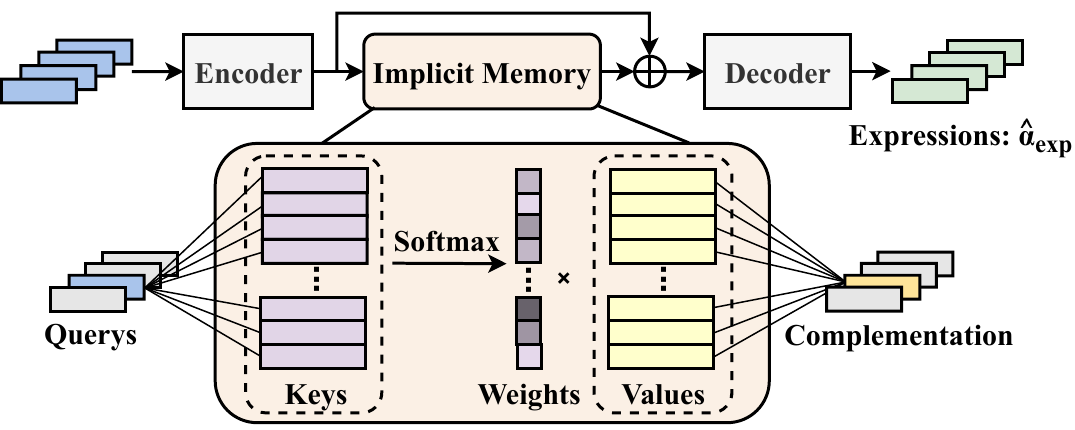}
  \caption{Our implicit memory is introduced between the encoder and decoder of the audio-to-expression model. The keys and values of the memory are both jointly learned with the audio-to-expression model.}
  \vspace{-4mm}
  \label{fig:a2e_model}
\end{figure}

Given an audio feature $\rmA \in \mathbb{R}^{T \times h_a}$ as input, the audio-to-expression model $f$ is expected to predict semantically-aligned expression coefficients $\boldsymbol{\hat{\alpha}}_{exp} \in \mathbb{R}^{T \times h_c}$ corresponding to the speech content, where $T$ is the number of frames, $h_a$ is the dimension of the audio feature~\cite{amodei2016deep}, $h_c$ is the dimension of the mouth-related expression coefficients~\cite{wood2021fake}. To address this sequence-to-sequence problem, we implement $f$ with a Transformer\cite{vaswani2017attention}-based architecture. As demonstrated in Sec.~\ref{sec:intro}, since there exists one-to-many mapping difficulty (\eg, one speech content may have multiple feasible expressions due to different habits and attitudes), directly predicting expression from the input audio feature alone is quite challenging. Therefore, considering the prediction of expression is semantically aligned to the input audio, we introduce the implicit memory to the audio-to-expression model $f$ to alleviate this one-to-many mapping.

\vspace{-3mm}
\paragraph{The implicit memory in the audio-to-expression model.}
In form of implicit memory, both the key set and the value set are randomly initialized at the beginning of the training, and are updated according to the backpropagation of the error signal during the training process.

As shown in Fig.~\ref{fig:a2e_model}, our audio-to-expression (\ie, a2e) model $f$ consists of an encoder $f_{enc}$ and a decoder $f_{dec}$. The implicit memory is introduced between the encoder and the decoder. Specifically, the feature extracted by the encoder is employed as the query $\rmQ^{a2e} = f_{enc}(\rmA)$. Then we take the inner product followed by a softmax function between the $\rmQ^{a2e}$ and the key set $\rmK^{a2e}$, obtaining a weighted representation of the value set $\rmV^{a2e}$. The attention result, which is served as the complemented information, is then added to the $\rmQ^{a2e}$ in an element-wise manner to form the input of the decoder $f_{dec}$. Formally:
\begin{equation}
    \label{equ:mem_a2e}
    \boldsymbol{\hat{\alpha}}_{exp} = f_{dec}(\rmQ^{a2e} \oplus \texttt{attn} (\rmQ^{a2e}, \rmK^{a2e}, \rmV^{a2e})),
    \vspace{-0.1cm}
\end{equation}
where $\oplus$ denotes the element-wise addition.
The size of $\rmK^{a2e}$ and $\rmV^{a2e}$ here are both $M \times h_a$, where $M$ stands for the number of keys and values.

\vspace{-3mm}
\paragraph{Loss functions.}
We adopt three loss functions to train the audio-to-expression model. 1) We minimize the $l_2$ distance between the predicted expression coefficients $\boldsymbol{\hat{\alpha}}_{exp}$ and the ground-truth one $\boldsymbol{\alpha}_{exp}$. 2) We minimize the distance between the predicted vertices $\hat{\rmO}_{m}$ and the ground-truth one $\rmO_{m}$ (see preliminaries for vertices acquirement). The size of vertices is $T \times h_v \times 3$, where $h_v$ denotes the number of vertices, and $3$ denotes the dimension of the coordinate. 3) To prevent each item in the memory to be similar to each other so as to increase the memory capacity, we also provide a regularization term on $\rmK^{a2e}$ and $\rmV^{a2e}$ during the training:
\begin{equation}
    \begin{aligned}
        L_{cof}^{a2e} &= \Vert \boldsymbol{\alpha}_{exp} - \boldsymbol{\hat{\alpha}}_{exp} \Vert_2, \\
        L_{vtx}^{a2e} &= \Vert \rmO_{m} - \hat{\rmO}_{m} \Vert_2, \\
        L_{reg}^{a2e} &= \frac{1}{M(M-1)}( \texttt{corr}(\rmK^{a2e}) + \texttt{corr}(\rmV^{a2e}) ), \\
        L^{a2e} &= \lambda_{cof} L_{cof}^{a2e} + \lambda_{vtx} L_{vtx}^{a2e} + \lambda_{reg} L_{reg}^{a2e},
    \end{aligned}
    \label{equ:a2e_loss}
\end{equation}
where $\texttt{corr} (\rmX) = \sum_i^M \sum_{j, j \neq i}^M \texttt{cos} (\rmX_i, \rmX_j)$, and $\texttt{cos} (\cdot, \cdot)$ denotes the cosine similarity between two vectors following previous works~\cite{casanova2022yourtts,xin2021cross,wu2022adaspeech}, $\lambda_{cof}$, $\lambda_{vtx}$ and $\lambda_{reg}$ are weights for each loss item. In total, we use $L^{a2e}$ to train the audio-to-expression model with the implicit memory.

\begin{figure}[t]
  \centering
  \includegraphics[width=1\linewidth]{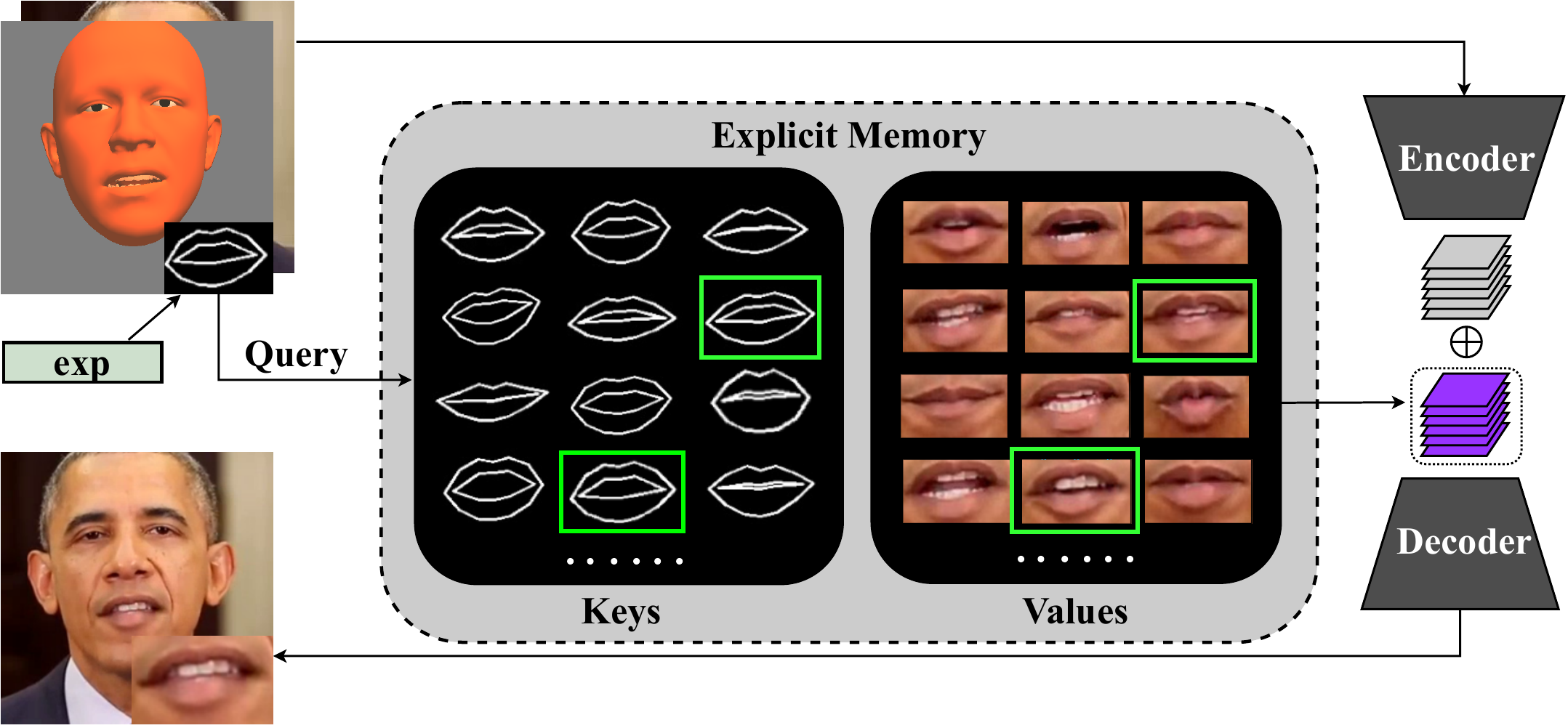}
  \caption{We construct the explicit memory by the vertices (\ie, the mouth shapes) and corresponding image patches, which can be retrieved by the neural-rendering model to complement the personalized visual details.}
  \vspace{-4mm}
  \label{fig:rendering_model}
\end{figure}

\subsection{Complementing Visual Appearance with Explicit Memory}
\label{sec:nr}
Given the predicted expression coefficients $\boldsymbol{\hat{\alpha}}_{exp}$, and the template video, the neural-rendering model $g$ is responsible for synthesizing the photo-realistic results.
As demonstrated in Sec.~\ref{sec:intro}, there also exists one-to-many mapping difficulty in the neural-rendering model. For example, one expression may have multiple feasible visual appearances due to different wrinkles and illuminations. Therefore, to complement the pixel-level details of the target person, an explicit memory is introduced to the neural-rendering model.

\vspace{-3mm}
\paragraph{The explicit memory in the neural-rendering model.}

Different from the implicit memory that learns the key set $\rmK$ and the value set $\rmV$ from training data automatically, the explicit memory constructs the key set $\rmK$ and the value set $\rmV$ directly from the data. For example, the key-value form can be formulated as representation-token pair in~\cite{khandelwal2021nearest}, text-image pair in~\cite{sheynin2022knn}, and vertex-image pair in our case.

As shown in Fig.~\ref{fig:rendering_model}, our neural-rendering (\ie, nr) model $g$ adopts an encoder-decoder architecture~\cite{ronneberger2015u}, where the explicit memory is introduced between the encoder $g_{enc}$ and the decoder $g_{dec}$. Specifically, we adopt the coordinates of vertices $\rmO_{m}$ as key set $\rmK^{nr}$ (see preliminaries for vertices acquirement), and its corresponding image of mouth region as the value set $\rmV^{nr}$. We build the explicit memory based on $\rmK^{nr}$ and $\rmV^{nr}$ for each target person with $N$ key-value pairs.
To make each item in the memory to be dissimilar so as to increase the capacity of the memory, we identify $N$ most dissimilar mouth shapes by calculating the Root-Mean-Square distance between mouth-related vertices. Then we use the $N$ vertex-image pairs to build the explicit memory. More details can be found in Sec.~\ref{sup:method} of supplementary materials.

For the predicted expression coefficients $\boldsymbol{\hat{\alpha}}_{exp}$, we obtain the pre-defined mouth-related vertices $\hat{\rmO}_{m}$ as elaborated in the preliminaries, and regard it as the query $\rmQ^{nr}$ to attend to the constructed explicit memory. The attention result, which includes substantial pixel-level details of the target person, is then merged with the feature $\rmF^{nr}$ extracted from the input of $g$ by the encoder $g_{enc}$ to generate the output:
\begin{equation}
    \label{equ:mem_nr}
    \hat{\rmI}_{RGB} = g_{dec}(\rmF^{nr} \oplus \texttt{attn} (\rmQ^{nr}, \rmK^{nr}, \rmV^{nr})),
    \vspace{-0.1cm}
\end{equation}
where $\hat{\rmI}_{RGB}$ denotes the final synthesized results.
Intuitively, by using the explicit memory, the closer the mouth-related vertices between $\rmQ^{nr}$ and $\rmK^{nr}$, the larger weight the corresponding mouth-related pixels will be returned as the complemented information for synthesis, thus making the prediction easier.

\vspace{-3mm}
\paragraph{Adaptation to new speakers.}

Since the explicit memory stores the person-specific visual appearance, when meeting a new speaker, the explicit memory is rebuilt from the talking video of the new speaker. In this way, our neural-rendering model can be flexibly adapted to the new speaker.

\vspace{-3mm}
\paragraph{Loss functions.}
We adopt two kinds of loss functions to optimize the neural-rendering model. 1) A reconstruction loss and a VGG perceptual loss~\cite{johnson2016perceptual} are employed to penalize the errors between the synthesized image and the ground-truth image. 2) To generate more realistic results, we employ a discriminator $d$ to verify whether the input is a natural image or a faked image produced by the neural-rendering model~\cite{goodfellow2014generative,thies2020neural}. Formally:
\begin{equation}
    \begin{aligned}
        L_{rec}^{nr} &= \Vert \rmI_{RGB} - \hat{\rmI}_{RGB} \Vert_2 + \texttt{VGG} (\rmI_{RGB}, \hat{\rmI}_{RGB}), \\
        L_{adv}^{d} &= - \mathbb{E} (\log d (\rmI_{RGB})) - \mathbb{E} (\log ( 1 - d (\hat{\rmI}_{RGB})), \\
        L_{adv}^{nr} &= \mathbb{E} (\log ( 1 - d (\hat{\rmI}_{RGB})), \\
        L^{nr} &= \lambda_{rec} L_{rec}^{nr} + \lambda_{adv} L_{adv}^{d} + \lambda_{adv} L_{adv}^{nr},
    \end{aligned}
    \label{equ:a2e_loss}
\end{equation}
where $\texttt{VGG} (\cdot, \cdot)$ denotes the VGG perceptual loss~\cite{johnson2016perceptual}, $\lambda_{rec}$ and $\lambda_{adv}$ are weights for each loss item. In total, we use $L^{nr}$ to train the neural-rendering model.

\subsection{Discussions}
\label{sec:discussion}

\paragraph{Why do we alleviate one-to-many mapping with memory?}

We note that one-to-many relationship exists in a wide range of problems~\cite{li2021audio2gestures,qian2021speech,wang2022low}, such as text-to-speech~\cite{tan2022naturalspeech}, machine translation~\cite{wang2018three}, image translation~\cite{shen2020one,kazemi2018unsupervised}, \etc. Among them, generative models (\eg, generative adversarial network~\cite{goodfellow2014generative}, normalizing flow~\cite{dinh2016density}, variational auto-encoder~\cite{kingma2013auto}) demonstrate great potential for the challenge of one-to-many mapping, with the intuition in that learning a distribution instead of a deterministic result. In our work, we also leverage the advance of the generative adversarial network, and further present the memory mechanism as an orthogonal approach to alleviating one-to-many mapping. The incorporation of memory implicates an insight: since predicting the missing information is difficult, why not construct storage to complement the information to the input? We answer this question in this paper, and demonstrate the effectiveness of memory for talking face generation in Sec.~\ref{sec:exp}.

\vspace{-3mm}
\paragraph{Why do we use two memory variants for two stages respectively?}

Basically, the audio-to-expression model and neural-rendering model play different roles in talking face generation. The audio-to-expression model is responsible for generating semantically-aligned expressions from the input audio, while the neural-rendering model synthesizes the pixel-level visual appearance according to the estimated expressions. Therefore, to make each prediction easier, we devise an implicit memory and an explicit memory to follow the sense of these two models. Intuitively, the implicit memory is to learn the prior from the training data automatically, then the learned prior can be used to complement the information that is missed by the input, yielding more realistic generation results. For neural rendering, where each mouth shape shares a similar visual appearance, we explicitly construct the expression-related prior from the video and directly retrieve the prior to complement the input. Our experiments in Sec.~\ref{sec:ablation} show the effectiveness of each choice.

\vspace{-1.5mm}
\section{Experiments}
\label{sec:exp}

\vspace{-1mm}
In this section, we compare our MemFace with state-of-the-art methods, and provide ablation studies. Since we alleviate the one-to-many mapping problem and make the prediction easier, we also adapt our model to new speakers with few adaption data and make comparisons. More results can be found in Sec.~\ref{sup:exp_results} of supplementary materials.

\subsection{Experimental Setups}
\label{sec:exp_setups}

\vspace{-1mm}
\paragraph{Datasets.}
Following previous works~\cite{chen2020talking,wang2021audio2head}, we utilize GRID dataset~\cite{cooke2006audio} (20 speakers, about four hours video) to train the audio-to-expression model and utilize Obama dataset~\cite{suwajanakorn2017synthesizing} (about 3 minutes video) to train the neural-rendering model.
For experiments on adaptation to new speakers, we randomly collect 10 speakers with various talking styles from HDTF dataset~\cite{zhang2021flow}. We use training data with different durations (15s and 30s) for each speaker.

\vspace{-4mm}
\paragraph{Objective metrics.}
For objective evaluation, following previous works~\cite{guo2021ad,park2022synctalkface,thies2020neural,shen2022learning}, 
we adopt \textbf{Sync-C} (SyncNet Confidence) and \textbf{Sync-D} (SyncNet Distance) to measure the lip-sync quality via SyncNet\cite{chung2016out}. Higher Sync-C or lower Sync-D indicates better lip-sync quality.
Since we can obtain vertices from the 3D face model (see Sec.~\ref{sec:nr} for details), we adopt \textbf{RMSE} (Root Mean Square Error) score between the ground-truth vertices and our predicted one to further evaluate the lip-sync quality. Lower RMSE indicates better lip-sync quality. We also adopt commonly used \textbf{LPIPS}\cite{zhang2018unreasonable} (Learned Perceptual Image Patch Similarity) score to evaluate the perceptual similarity between ground-truth images and the generated ones. Lower LPIPS indicates better perceptual quality.

\vspace{-4mm}
\paragraph{Subjective metrics.}
For subjective evaluation, 20 experienced participants are invited and each participant needs to make around $104$ judgments in total. We make detailed guidelines and example videos to ensure consistent grading criteria.
We use MOS (Mean Opinion Score) and \textbf{CMOS} (Comparison Mean Opinion Score) as our metrics. MOS contains subjective scores on lip-sync quality (\textbf{Lip-sync}), rendering quality (\textbf{Render}), and overall quality (\textbf{Overall}).
Higher MOS or CMOS indicates better subjective quality. More details for subjective evaluation are provided in Sec.~\ref{sup:metric_detail} of supplementary materials.

\begin{figure}[t]
  \centering
  \includegraphics[width=0.99\linewidth]{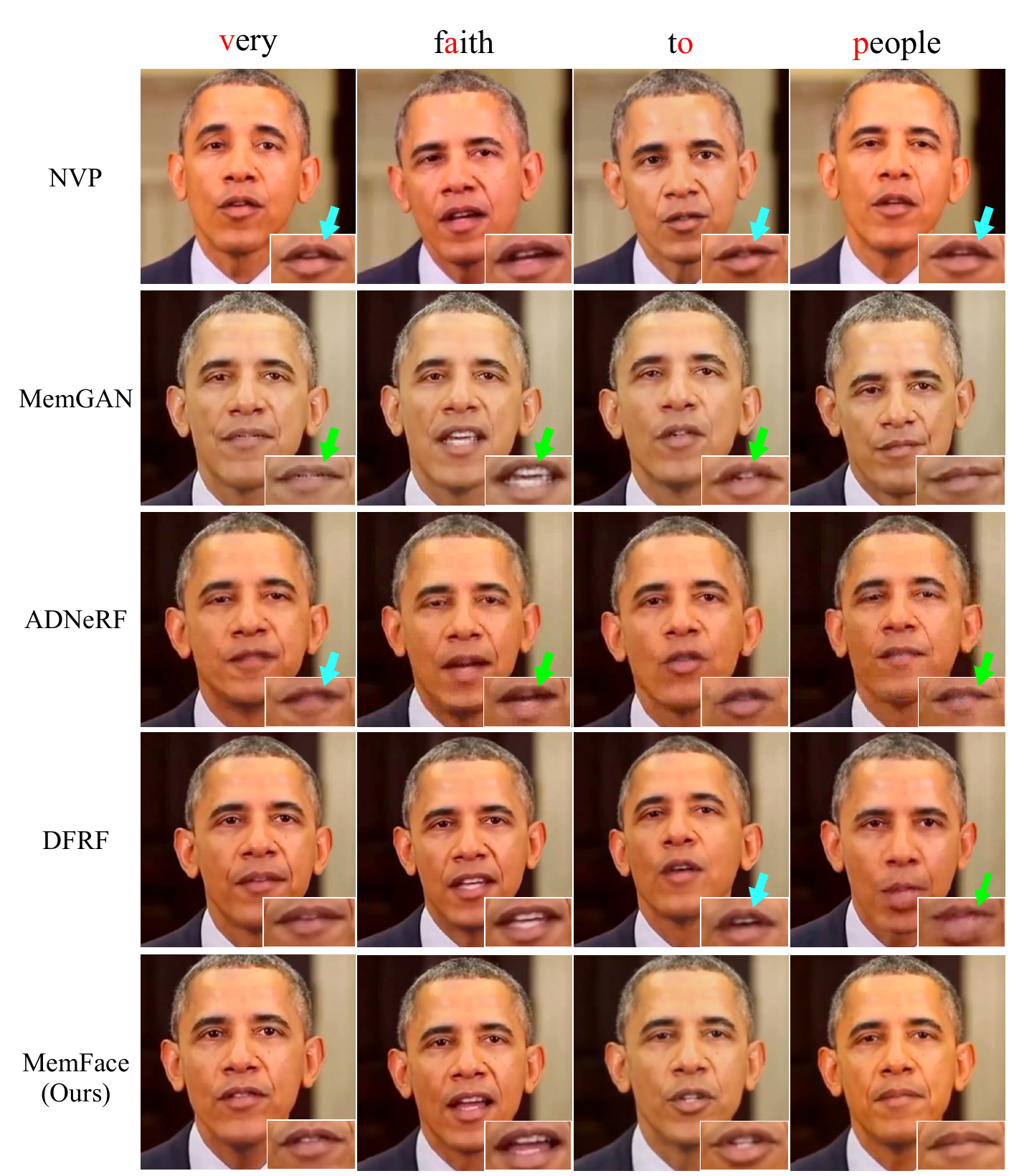}
  \vspace{-1mm}
  \caption{Qualitative comparison with the state-of-the-art methods (NVP~\cite{thies2020neural}, MemGAN~\cite{yi2020audio}, ADNeRF\cite{guo2021ad} and DFRF\cite{shen2022learning}) on Obama dataset. The blue and green arrows indicate inferior lip-sync and rendering quality respectively. It shows that our MemFace achieves higher lip-sync and rendering quality.}
  \label{fig:general_results_comp}
\end{figure}

\begin{table}[t]
\setlength\tabcolsep{3pt}
\begin{center}
\footnotesize
\caption{Quantitative comparison with the state-of-the-art methods on Obama dataset. Our MemFace achieves the best subjective and objective quality by a large margin. We highlight the best and the second-best number in bold and underline respectively.}
    \begin{tabular}{lccccc}
    \toprule[1.5pt]
    \multirow{2}{*}{Method} & \multicolumn{3}{c}{Subjective Evaluation} & \multicolumn{2}{c}{Objective Evaluation} \\
    \cmidrule(l){2-4}
    \cmidrule(l){5-6}

     & Lip-sync$\uparrow$  & Render$\uparrow$ & Overall$\uparrow$ & Sync-C $\uparrow$ & Sync-D $\downarrow$  \\
    \midrule
    NVP\cite{thies2020neural}  & 3.399 & 3.173   & 3.364  &  4.428 & 9.788 \\
    LipSync3D\cite{lahiri2021lipsync3d} & 3.355  & 3.099  & 3.109   &  5.466  & \underline{8.495} \\
    MemGAN\cite{yi2020audio}& 2.667 & 2.833 & 2.417 & 4.017 & 10.05 \\
    ADNeRF\cite{guo2021ad} &  \underline{3.691} & \underline{3.900} & \underline{3.700} & \underline{5.723} & 9.164   \\
    DFRF\cite{shen2022learning} &  3.155 & 3.045 & 3.109 & 4.260 & 10.582  \\
    \midrule
    MemFace (Ours) &  \textbf{4.381} & \textbf{3.909}  & \textbf{4.318} & \textbf{6.022} & \textbf{8.419}  \\
    \bottomrule[1.5pt]
    \end{tabular}
    \vspace{-4mm}
    \label{tab:general_comp}
\end{center}
\end{table}

\begin{table*}[tb]
\setlength\tabcolsep{8pt}
\begin{center}
\footnotesize
\caption{We adapt MemFace to new speakers with few adaption data, and compare it to the state-of-the-art methods with the same setting.}
    \vspace{-1mm}
    \begin{tabular}{clccccc}
    \toprule[1.5pt]
    \multirow{2}{*}{Adaption Set} & \multirow{2}{*}{Method} &
    \multicolumn{3}{c}{Subjective Evaluation} & \multicolumn{2}{c}{Objective Evaluation} \\
    \cmidrule(l){3-5}
    \cmidrule(l){6-7}
    & & Lip-sync $\uparrow$ & Render$\uparrow$ & Overall $\uparrow$  & Sync-C $\uparrow$ & Sync-D $\downarrow$ \\
    \midrule
    
    \multirow{4}{*}{15s} & AD-NeRF\cite{guo2021ad}  & 1.9419 & 2.4347 & 2.1883 &1.606 & 12.259\\
    & DFRF\cite{shen2022learning} & \underline{2.2753} & \underline{2.8116} & \underline{2.2028} & 2.360 & 11.534\\
    & MemGAN\cite{yi2020audio}  & 2.2217 & 2.5726 & 2.1022 & \underline{4.058} & \underline{10.189} \\
    \midrule
    & MemFace (Ours) & \textbf{4.1288} & \textbf{3.8231} & \textbf{3.9842} & \textbf{5.385} & \textbf{9.003}\\ 
    \toprule[1.5pt]

    \multirow{4}{*}{30s} & AD-NeRF\cite{guo2021ad}  & 2.2896 & 2.4057 & 2.1304 & 1.028 & 13.445 \\
    & DFRF\cite{shen2022learning} & 2.8116 & \underline{2.7536} & \underline{2.5507} & 3.607 & 10.908 \\
    & MemGAN\cite{yi2020audio}  & \underline{2.4058} & 2.3187 & 1.6811 & \underline{4.410} & \underline{9.998}\\
    \midrule
    & MemFace (Ours) & \textbf{4.4405} & \textbf{3.9854} & \textbf{4.1737} & \textbf{6.058}  & \textbf{8.431}\\ 
    \bottomrule[1.5pt]
    \end{tabular}
    \vspace{-6mm}
    \label{tab:adaptation_comp}
\end{center}
\end{table*}

\vspace{-4mm}
\paragraph{Implementation details.}
For the dimension of the hidden states, we use $h_a=64$~\cite{amodei2016deep}, $h_c=85$~\cite{wood2021fake} and $h_v=69$~\cite{wood2021fake} in our experiments.
In this paper, we adopt the image resolution of $256\times256$.
$\lambda_{cof}$, $\lambda_{vtx}$, $\lambda_{reg}$, $\lambda_{rec}$ and $\lambda_{adv}$ are empirically set to $1$, $1$, $0.1$, $20$ and $1$ respectively. Unless it is stated, we use $M=1000$ and $N=300$.
We adopt the Adam~\cite{kingma2014adam} optimizer with a learning rate of $1e-4$.
More details are provided in Sec.~\ref{sup:exp_detail} of supplementary materials.

\vspace{-2mm}
\subsection{Comparisons with Previous Models}

\vspace{-1mm}
\paragraph{Our MemFace outperforms the state-of-the-art methods in both objective and subjective evaluation.}

With the same input audio, we compare our synthesized results with the following works: NVP~\cite{thies2020neural}, LipSync3D~\cite{lahiri2021lipsync3d}, MemGAN~\cite{yi2020audio}, ADNeRF\cite{guo2021ad} and DFRF\cite{shen2022learning}.
In Fig.~\ref{fig:general_results_comp}, we present the synthesized results of different methods when saying the same syllable. 
It demonstrates that, compared to the state-of-the-art methods, our synthesized results in the last row exhibit more accurate lip-sync and more satisfactory rendering quality.
Quantitative results are shown in Tab.~\ref{tab:general_comp}, where
our method generally outperforms the comparison ones in both subjective and objective evaluation. 

\begin{figure}[t]
  \centering
  \includegraphics[width=1\linewidth]{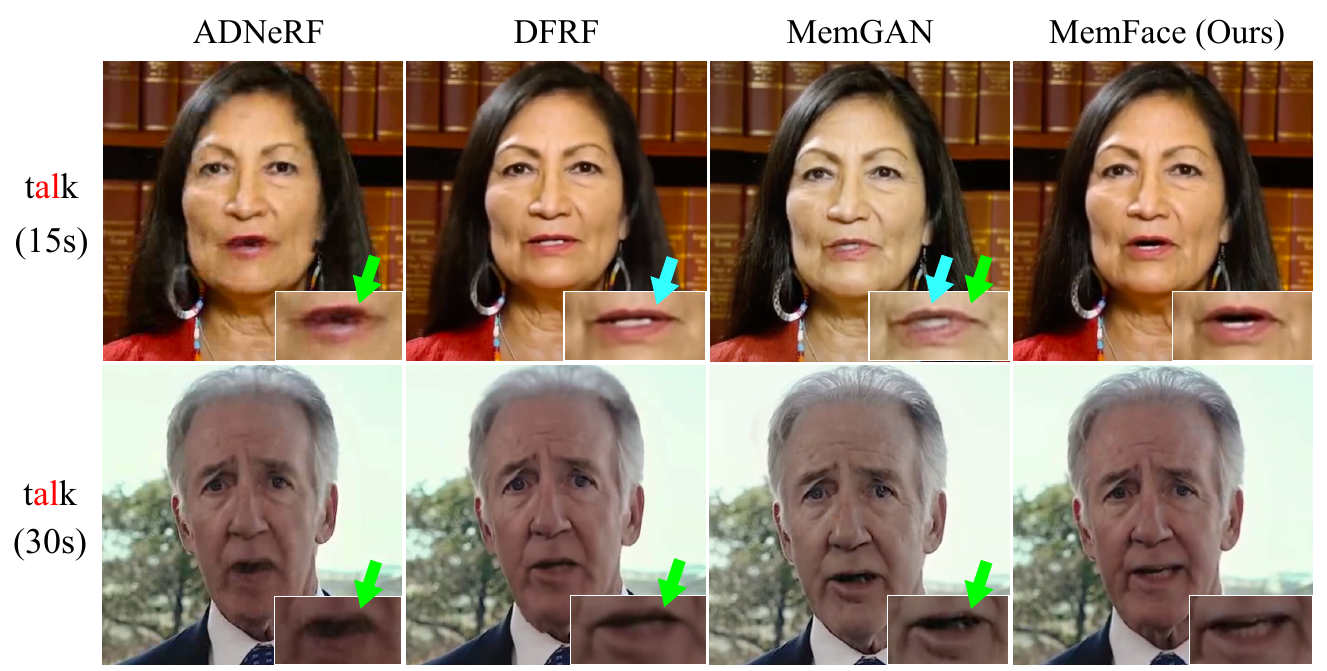}
  \caption{Qualitative comparison with the state-of-the-art methods (ADNeRF\cite{guo2021ad}, DFRF\cite{shen2022learning} and MemGAN~\cite{yi2020audio}) when adapting the model to new speakers with few adaption data. The blue and green arrows indicate inferior lip-sync and rendering quality respectively.}
  \label{fig:ft_comp}
\vspace{-4mm}
\end{figure}

\vspace{-4mm}
\paragraph{Our MemFace outperforms the state-of-the-art methods when adapting the model to new speakers with few adaption data.}
As discussed in Sec.~\ref{sec:discussion}, to alleviate the one-to-many mapping, we ease the prediction by complementing the missing information with memories. To verify that the memories make the prediction easier, we conduct experiments on adapting our model to new speakers with few adaption data (\ie, 15s and 30s training videos from HDTF dataset~\cite{zhang2021flow}). 
Specifically, for both our MemFace and the state-of-the-art methods~\cite{guo2021ad,shen2022learning,yi2020audio}, we first pre-train the models on the Obama dataset. Then, we build the explicit memory for each identity from the short adaption video. Finally, we fine-tune both the audio-to-expression model and neural-rendering model on the short adaption videos based on the pre-trained model parameters. As shown in Table.~\ref{tab:adaptation_comp}, our method exhibits an obviously better performance in all aspects under different scenarios. We also present some qualitative results in Fig.~\ref{fig:ft_comp} to make an intuitive comparison.
It can be seen that, the baseline methods tend to generate blurry images~\cite{guo2021ad,shen2022learning} or inaccurate lip motion~\cite{shen2022learning,yi2020audio}. In contrast, our adaptation results retain more texture details such as the teeth, and achieve a satisfactory lip-sync quality for different talking styles.

\vspace{-1mm}
\subsection{Ablation Studies}
\label{sec:ablation}

\vspace{-1mm}
In this section, we verify the choices of implicit and explicit memory for each stage, and provide ablation studies on the hyper-parameters of memory (\ie, the number of key-value pairs). More ablation studies are provided in Sec.~\ref{sup:exp_results} of supplementary materials.

\vspace{-3mm}
\paragraph{The implicit memory is more suitable for audio-to-expression model.}

\begin{figure}[t]
  \centering
  \includegraphics[width=1\linewidth]{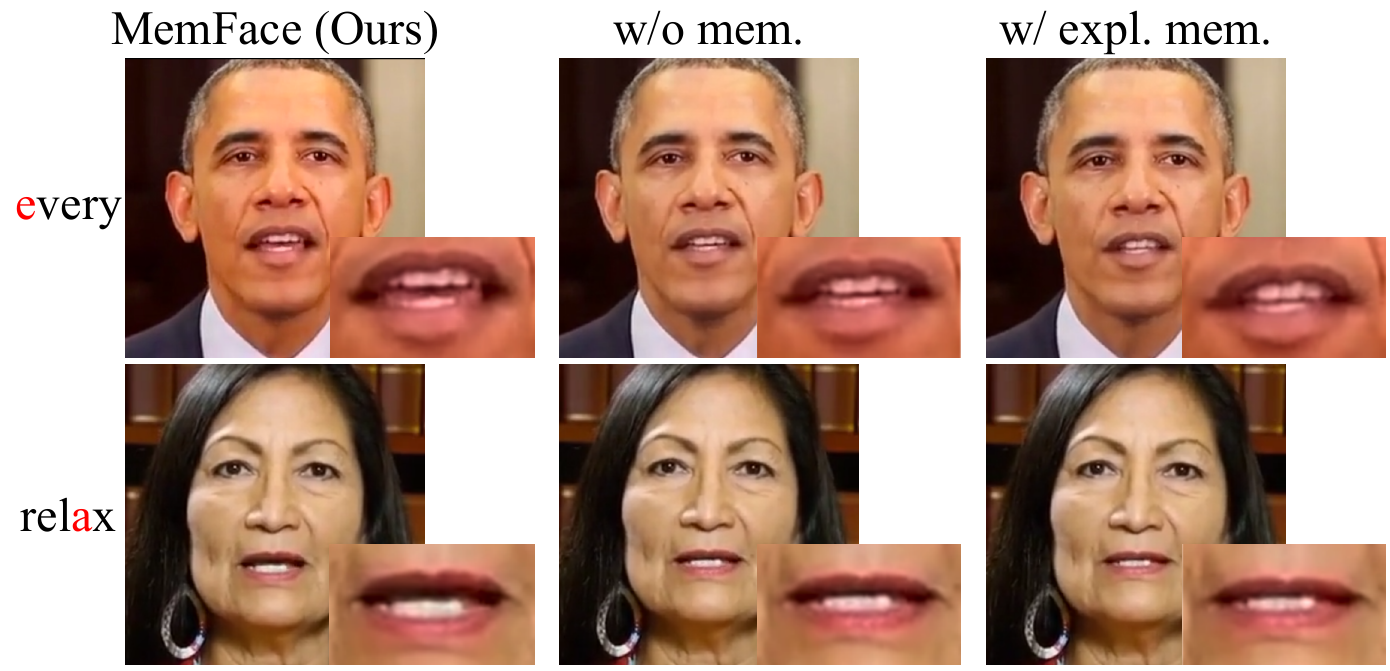}
  \caption{Ablation results of the implicit memory in audio-to-expression model. We use the same neural-rendering models with explicit memory for different settings. It can be observed that the audio-to-expression model with the implicit memory (MemFace) generates higher lip-sync quality.}
  \vspace{-2mm}
  \label{fig:a2e_abla}
\end{figure}

We introduce an implicit memory to the audio-to-expression model. To verify which memory is suitable for the audio-to-expression model, we remove the implicit memory (w/o mem.) or replace it with explicit memory (w/ expl. mem.), where the audio features $\rmA$ and expression coefficients $\boldsymbol{\hat{\alpha}}_{exp}$ are regarded as keys and values respectively. We train the audio-to-expression model with the GRID dataset and then adapt the model to the 15s HDTF dataset to study the memory choice. In Tab.~\ref{tab:abla_a2e_memory} and Fig.~\ref{fig:a2e_abla}, it can be observed that compared with both settings (\ie, w/o mem. \& w/ expl. mem.), our scheme with implicit memory achieves the better objective and subjective quality, which verifies that the implicit memory is more suitable for the audio-to-expression model. It also demonstrates that replacing the implicit memory with the explicit memory even leads to inferior quality than removing the implicit memory for the audio-to-expression model. We argue that the prediction of expression is semantically aligned with the input audio, and depends on the audio context. Therefore, directly retrieving the corresponding expression leads to more noisy expressions, thus damaging the performance.

\begin{table}[t]
\begin{center}
\footnotesize
	\caption{Abalation studies on the implicit memory in the audio-to-expression model. The CMOS criterion represents lip-sync quality here. See Sec.~\ref{sec:ablation} for detailed descriptions. 
	}
 \vspace{-1.5mm}
	\label{tab:abla_a2e_memory}
	\begin{tabular}{lcccc}
		\toprule[1.5pt]
		 \multirow{2}{*}{Method} & \multicolumn{2}{c}{GRID} & \multicolumn{2}{c}{15s HDTF} \\
		 \cmidrule(l){2-3}
		 \cmidrule(l){4-5}
		 & RMSE$\downarrow$ & CMOS$\uparrow$ & RMSE$\downarrow$  & CMOS$\uparrow$ \\
		\midrule
		MemFace    & \textbf{0.0826}  & \textbf{0}  & \textbf{0.1343}  & \textbf{0} \\
		w/o mem.         &  0.0834 & -0.1090   &  0.1448  & -0.1136  \\
		w/ expl. mem.  & 0.0846 &  -0.2363   &  0.1398  & -0.1136   \\
		\bottomrule[1.5pt]
	\end{tabular}
\end{center}
\vspace{-3mm}
\end{table}

\begin{table}[t]
\begin{center}
\footnotesize
	\caption{Abalation study on the explicit memory in the neural-rendering model. The CMOS criterion represents rendering quality here. Sec.~\ref{sec:ablation} for detailed descriptions. 
	}
     \vspace{-1.5mm}
	\label{tab:abla_rendering_memory}
	\begin{tabular}{lcccc}
		\toprule[1.5pt]
		 \multirow{2}{*}{Method} & \multicolumn{2}{c}{Obama} & \multicolumn{2}{c}{15s HDTF} \\
		 \cmidrule(l){2-3}
		 \cmidrule(l){4-5}
		 & LPIPS$\downarrow$ & CMOS$\uparrow$ & LPIPS$\downarrow$  & CMOS$\uparrow$ \\
		\midrule
		MemFace    & \textbf{0.0207}  & \textbf{0}  & \textbf{0.0218}  & \textbf{0} \\
		w/o mem.         &  0.0248 & -0.0454   &  0.0254  & -0.0545  \\
		w/ impl. mem.  & 0.0245 &  -0.0281   &  0.0264  & -0.0363   \\
		\bottomrule[1.5pt]
	\end{tabular}
\end{center}
\vspace{-3mm}
\end{table}

\begin{table}[t]
\caption{Ablation studies on the number of key-value pairs in two memories ($M$ and $N$). Param. indicates the number of parameters of the audio-to-expression model. We train the neural-rendering model using Obama (LPIPS-1) and 15s HDTF (LPIPS-2, $375$ frames) data to study the effect of the memory capacity.}
\vspace{-1mm}
\footnotesize
    \begin{subtable}[t]{0.495\linewidth}
    \caption{The implicit memory.}
    \centering
    \setlength\tabcolsep{6pt}
        \begin{tabular}[t]{@{}l|cc@{}}
    	    \toprule[1.5pt]
    		$M$  & Param. &  RMSE$\downarrow$  \\
    		\midrule
    		500         & 177k   & 0.0828      \\
    		\textbf{1000}  & \textbf{240k}    & \textbf{0.0826 }    \\
    		1500               & 305k    & 0.0826    \\
    		2000    &  369k & 0.0825 \\
    		\bottomrule[1.5pt]
    	\end{tabular}
    \end{subtable}
    \begin{subtable}[t]{0.495\linewidth}
    \caption{The explicit memory.}
    \centering
    \setlength\tabcolsep{6pt}
        \begin{tabular}[t]{@{}l|cc@{}}
    	    \toprule[1.5pt]
    		$N$  & LPIPS-1$\downarrow$ &  LPIPS-2$\downarrow$  \\
    		\midrule
    		100         & 0.0218   & 0.0237      \\
    		200       & 0.0215    & 0.0224    \\
    		\textbf{300}  & \textbf{0.0207}    & \textbf{0.0218}    \\
    		500    &  0.0205 & - \\
    		\bottomrule[1.5pt]
    	\end{tabular}
    \end{subtable}
    \label{tab:array}
\vspace{-3mm}
\end{table}

\vspace{-3mm}
\paragraph{The explicit memory is more suitable for the neural-rendering model.}

\begin{figure}[t]
  \centering
  \includegraphics[width=1\linewidth]{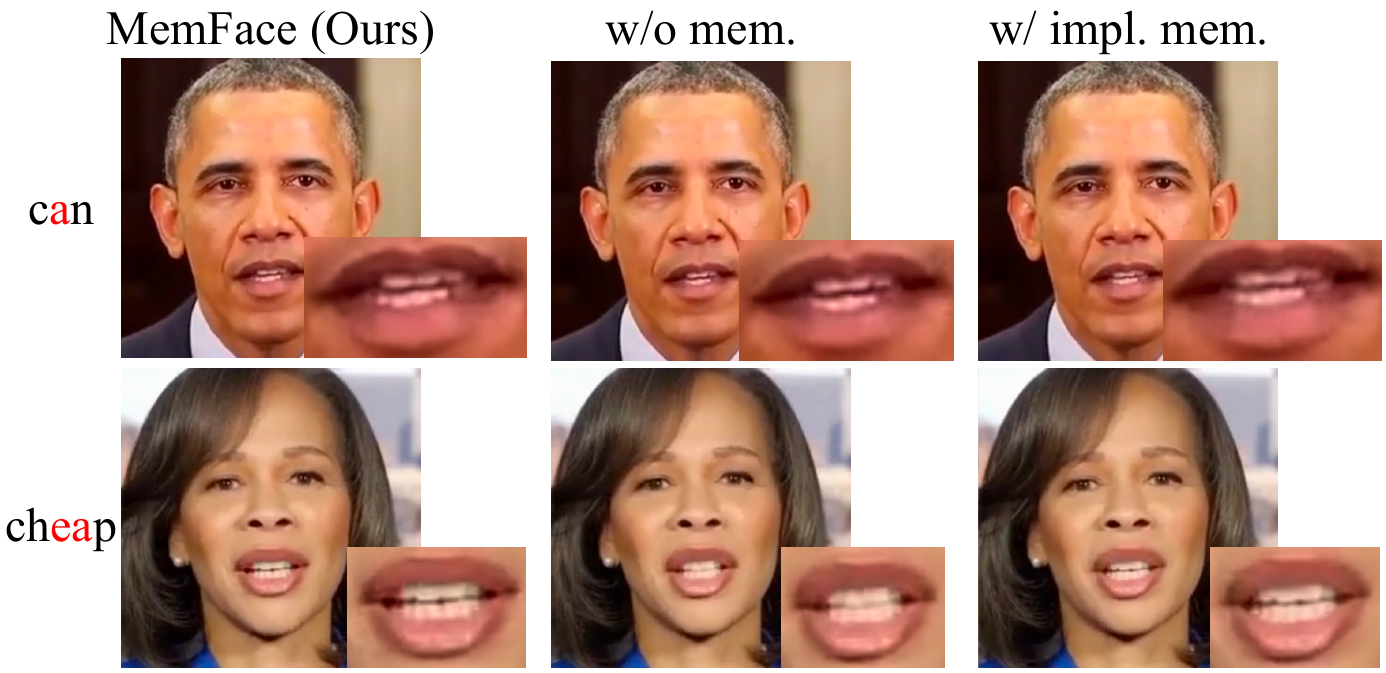}
  \caption{Ablation results of the explicit memory in the neural-rendering model. We use the same audio-to-expression models with the implicit memory for different settings. It can be observed that the neural-rendering model with the explicit memory (MemFace) synthesizes more realistic visual appearances.}
  \vspace{-2mm}
  \label{fig:rendering_abla}
\end{figure}

In the neural-rendering model, we use explicit memory to complement the personalized visual details. We also verify which memory is suitable for the neural-rendering model in Tab.~\ref{tab:abla_rendering_memory} and Fig.~\ref{fig:rendering_abla}, in which we remove the explicit memory (w/o mem.) or replace it with implicit memory (w/ impl. mem.). The keys and values of this implicit memory are randomly initialized and jointly trained with the neural-rendering model. We train the neural-rendering model with the Obama dataset and then adapt the model to the 15s HDTF dataset. It can be observed that our scheme with explicit memory achieves the best objective and subjective quality. It also demonstrates that although employing an implicit memory (w/ impl. mem.) in the neural-rendering model can complement the missing information to some extent, our explicit memory (MemFace) provide more visual details and yield better synthesis quality.


\vspace{-3mm}
\paragraph{Ablation studies on the number of key-value pairs in two memories.}
To study the effect of memory capacity, we vary the number of key-value pairs in two memories (\ie, $M$ and $N$ for the implicit and explicit memory respectively). The results shown in Tab.~\ref{tab:array} indicate that $M=1000$ and $N=300$ achieve the best objective quality, which also shows that our model does not depend on larger memory capacity.

\vspace{-1mm}
\section{Conclusion}
\label{sec:conclusion}

   In this work, we improve both the lip-sync and rendering quality of talking face generation by alleviating the one-to-many mapping challenge with memories. Our MemFace incorporates an implicit memory and an explicit memory into the audio-to-expression model and the neural-rendering model respectively. The experiments verify the effectiveness of the two memories across multiple scenarios.

   For future works, it is worth applying our idea to other one-to-many mapping tasks, such as text-to-image generation, and image translation. We will also investigate better ways to alleviate one-to-many mapping difficulty.

   \paragraph{Ethical consideration.} Our MemFace is developed only for research purposes. It should not be abused for fake face synthesis.

{\small
\bibliographystyle{ieee_fullname}
\bibliography{egbib}
}

\appendix
\clearpage

In this supplementary, we provide more details and results to make our paper more comprehensive. Specifically, we present more method details in Sec.~\ref{sup:method}, more experimental details in Sec.~\ref{sup:exp_detail}, and more experimental results in Sec.~\ref{sup:exp_results}.

\section{Details of Method}
\label{sup:method}
\subsection{Construction of the explicit memory}
As illustrated in Sec.~\ref{sec:nr} of the main paper, to construct the explicit memory in the neural-rendering model, we identify $N$ most dissimilar mouth shapes by calculating the Root-Mean-Square distance between mouth-related vertices, and then use the $N$ vertex-image pairs to build the explicit memory. 
Obviously, exploring all possible cases is almost infeasible due to heavy computation.
So we adopt a manually-designed algorithm to construct the explicit memory and the detailed steps are shown in Alg.~\ref{alg:expl_mem}.

\begin{algorithm}[H]
    \caption{Construction of the explicit memory.}
    \label{alg:expl_mem}
    \KwIn{All vertex-image pairs ($\rmK_{all}^{nr}$ and $\rmV_{all}^{nr}$) in dataset.}
    \KwOut{$N$ vertex-image pairs ($\rmK^{nr}$ and $\rmV^{nr}$).}
    
    \begin{algorithmic}[1]
    \STATE Set $N$ and randomly select $N$ vertex-image pairs from $\rmK_{all}^{nr}$ and $\rmV_{all}^{nr}$ to initialize $\rmK^{nr}$ and $\rmV^{nr}$; 
    \STATE $(k_{m1},k_{m2},D_{min}) \leftarrow f(\rmK^{nr})$, where $f$ is a function identifying the most similar two mouth shapes $(k_{m1}, k_{m2}) \in \rmK^{nr}$ according to the corresponding Root-Mean-Square (RMS) distance $D_{min}$;
    \FOR{($k_{tmp}$, $v_{tmp}$) in ($\rmK_{all}^{nr}$, $\rmV_{all}^{nr}$)}
        \STATE $\rmK_{tmp1}^{nr} \leftarrow \rmK^{nr}$, replace $k_{m1} \in \rmK_{tmp1}^{nr}$ with $k_{tmp}$, $(..., D_{min1}) \leftarrow f(\rmK_{tmp1}^{nr})$; 
        \STATE $\rmK_{tmp2}^{nr} \leftarrow \rmK^{nr}$, replace $k_{m2} \in \rmK_{tmp2}^{nr}$ with $k_{tmp}$, $(..., D_{min2}) \leftarrow f(\rmK_{tmp2}^{nr})$;
        \IF{$\max(D_{min1}, D_{min2}) > D_{min}$}
        \IF{$D_{min1} > D_{min2}$}
        \STATE Replace ($k_{m1} \in \rmK^{nr}$,$v_{m1} \in \rmV^{nr}$) with ($k_{tmp}$, $v_{tmp}$);
        \ELSE
        \STATE Replace ($k_{m2} \in \rmK^{nr}$,$v_{m2} \in \rmV^{nr}$) with ($k_{tmp}$, $v_{tmp}$);
        \ENDIF
        \STATE $(k_{m1},k_{m2},D_{min}) \leftarrow f(\rmK^{nr})$;
        \ENDIF 
    \ENDFOR
    \STATE Return $\rmK^{nr}$ and $\rmV^{nr}$ as the $N$ vertex-image pairs to construct the explicit memory.
    \end{algorithmic}
\end{algorithm}

\section{Details of Experiments}
\label{sup:exp_detail}
\subsection{Dataset details}
\label{sup:dataset_detail}
\paragraph{Adaptation dataset.}
As described in Sec.~\ref{sec:exp_setups} of the main paper, we randomly collect 10 speakers with various talking styles from HDTF dataset~\cite{zhang2021flow}, and conduct two sets of experiments using adaptation data with different durations (15s and 30s). 
We additionally conduct experiments with a 60s adaptation set for cases with intense head movements, and the results are shown in Sec.~\ref{sup:supp_more_exp_results}.

\subsection{Training details}
We adopt Adam~\cite{kingma2014adam} optimizer with a learning rate of $1$e$-4$ in the training stage. For adaptation experiments, we adapt the audio-to-expression model with a learning rate of $5$e$-6$ ($200$ epochs) and the neural-rendering model with a learning rate of $1$e$-4$ ($50$ epochs). 
To make the training easier, for the implicit memory, we update model parameters and memory alternately in the first half of the training. 

\begin{table*}[t]
\setlength\tabcolsep{8pt}
\begin{center}
\small
\caption{More adaptation results.}
    \vspace{-2mm}
    \begin{tabular}{clccccc}
    \toprule[1.5pt]
    \multirow{2}{*}{Adaption Set} & \multirow{2}{*}{Method} &
    \multicolumn{3}{c}{Subjective Evaluation} & \multicolumn{2}{c}{Objective Evaluation} \\
    \cmidrule(l){3-5}
    \cmidrule(l){6-7}
    & & Lip-sync $\uparrow$ & Render$\uparrow$ & Overall $\uparrow$  & Sync-C $\uparrow$ & Sync-D $\downarrow$ \\
    \midrule
    \multirow{4}{*}{60s} & AD-NeRF\cite{guo2021ad}  & \underline{2.4347} & 2.2028 & 2.3477 & 1.677 & 13.049\\
    & DFRF\cite{shen2022learning} & 2.1738 & \underline{2.4637} & \underline{2.3767} & \underline{2.235} & \underline{12.170}\\
    & MemGAN\cite{yi2020audio}  & 1.6790 & 2.3187 & 1.7937 & 1.710& 12.872\\
    \midrule
    & MemFace (Ours) & \textbf{4.2746} & \textbf{3.9853}  & \textbf{4.2014} & \textbf{5.643} & \textbf{8.874} \\
    \bottomrule[1.5pt]
    \end{tabular}
    \vspace{-3mm}
    \label{tab:adaptation_60s}
\end{center}
\end{table*}

\begin{table}[t]
\caption{Ablation study on the size of the implicit memory. The number of key-value pairs is 1000, $D$ means the dimension of keys and values. Param. indicates the number of parameters of the audio-to-expression model. We perform training with GRID~\cite{cooke2006audio} data (RMSE-1) and perform adaptation with HDTF~\cite{zhang2021flow} data (RMSE-2) to study the effect of $D$.}
\vspace{-1mm}
\small
\centering
\label{tab:abla_size}
    \begin{tabular}[t]{@{}l|ccc@{}}
	    \toprule[1.5pt]
		$D$  & Param. & RMSE-1$\downarrow$ &  RMSE-2$\downarrow$  \\
		\midrule
		32  & 178k  & 0.0832  &  0.1341  \\
		\textbf{64}  &  \textbf{240k}   & \textbf{ 0.0826} &  \textbf{0.1343}  \\
		96  &  314k   &   0.0824  &  0.1352   \\
		128  &  382k  &  0.0823  &  0.1357  \\
		\bottomrule[1.5pt]
	\end{tabular}
 \vspace{-3mm}
\end{table}

\begin{figure*}[t]
  \centering
  \includegraphics[width=1\linewidth]{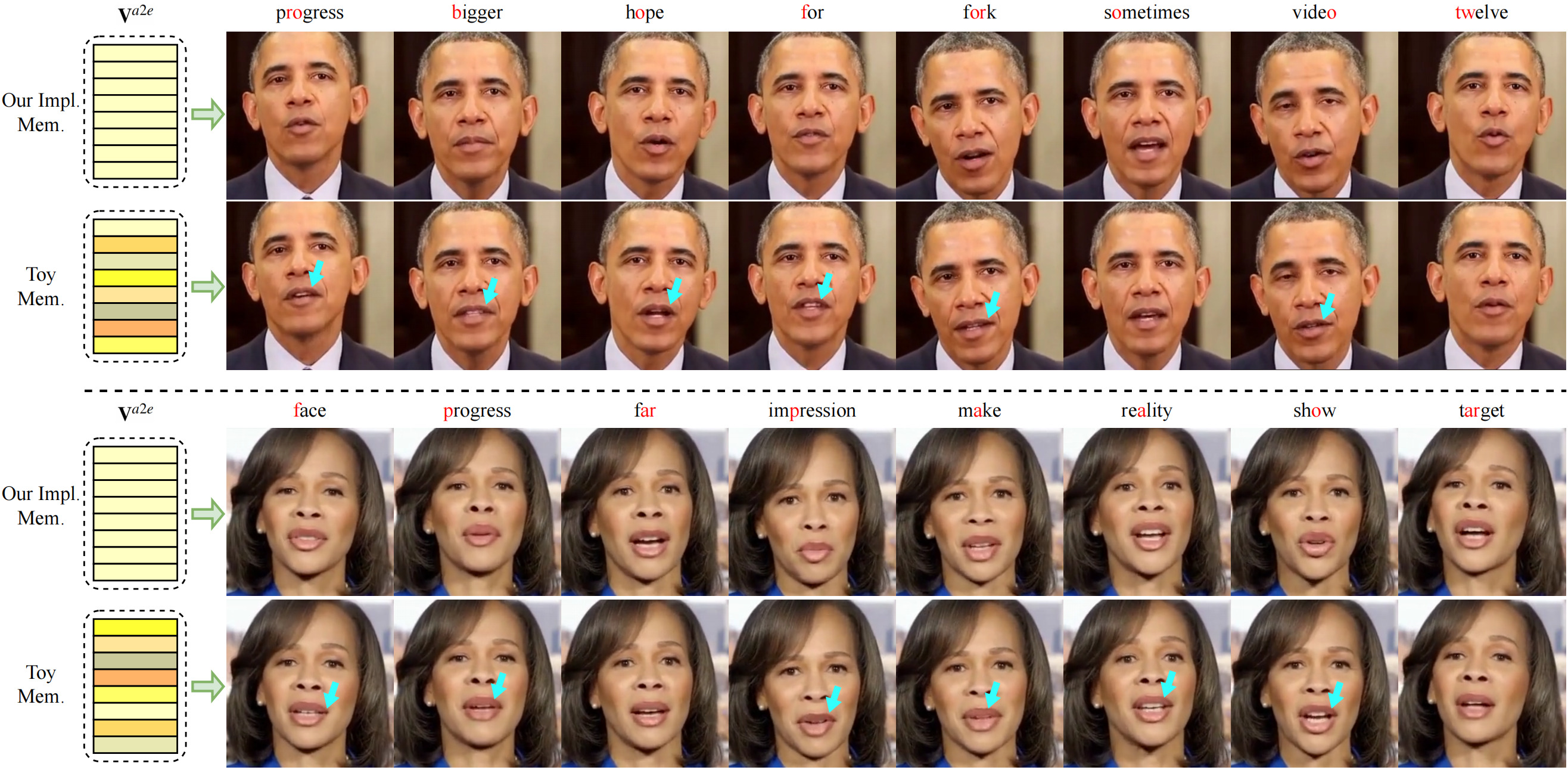}
  \caption{Set the implicit memory to random values. (upper half: train the audio-to-expression model on GRID~\cite{cooke2006audio}, lower half: adaption on HDTF~\cite{zhang2021flow}). The blue arrows indicate obvious inferior lip-sync quality.}
  \vspace{-3mm}
  \label{fig:supp_toy-a2e}
\end{figure*}

\begin{figure*}[t]
  \centering
  \includegraphics[width=1\linewidth]{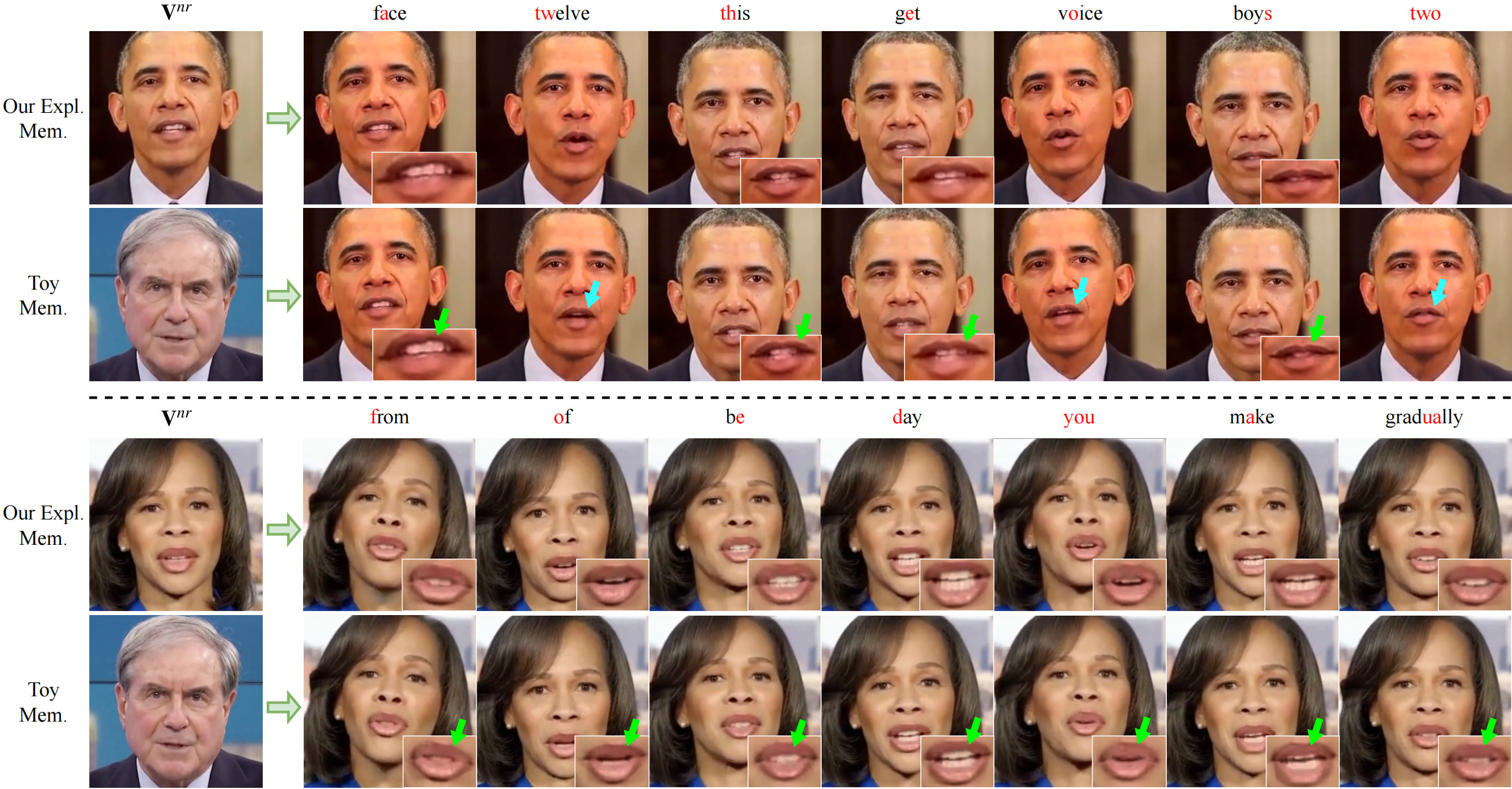}
  \caption{Set the explicit memory to the visual appearance of another person. (upper half: train the neural-rendering model on Obama~\cite{suwajanakorn2017synthesizing}, lower half: adaptation on HDTF~\cite{zhang2021flow}). The blue and green arrows indicate inferior lip-sync and rendering quality respectively.}
  \vspace{-0.5cm}
  \label{fig:supp_toy-render}
\end{figure*}

\subsection{Metrics}
\label{sup:metric_detail}
\paragraph{Details of subjective evaluation.}
For subjective evaluation, 20 experienced users are invited to participate. 
We use MOS (Mean Opinion Score) and CMOS (Comparison Mean Opinion Score) as our metrics in comparison experiments and ablation studies respectively. 
In MOS experiments, we present one video at a time and ask the users to rate the presented video at five grades in lip-sync, rendering, and overall quality respectively. 
Different grade choices in MOS are: Very Good (5), Good (4), Average (3), Poor (2), Very Poor (1). 
In CMOS experiments,
we present two videos at a time and ask the users to make
comparisons in lip-sync or rendering quality. 
Different grade choices in CMOS are: left video is obviously better (2), left video is a little better (1), indistinguishable (0), right video
is a little better (-1), right video is obviously better (-2). 

We conduct the online evaluations with well-designed Google questionnaires.
A detailed guideline with example videos is provided at the beginning of the questionnaires to ensure a consistent grading criterion.

\vspace{-3mm}
\section{More Experimental Results}
\label{sup:exp_results}
\subsection{More ablation studies}
\label{sup:supp_more_abla_studies}

\paragraph{Dimension of the implicit memory.}
In Sec.~\ref{sec:ablation} of the main paper, we conduct ablation studies on the number of key-value pairs in the implicit memory. As a supplement, we also explore the dimension selection of keys and values in the implicit memory. In Tab.~\ref{tab:abla_size}, we set the dimension of keys and values to $32$, $64$ (our adoption), $96$ and $128$ respectively and show the corresponding performance. From Tab.~\ref{tab:abla_size}, with the increase of dimensions, the prediction performance on GRID~\cite{cooke2006audio} dataset (RMSE-1) increases while the performance on HDTF~\cite{zhang2021flow} adaptation set (RMSE-2) decreases which may be caused by an overfitting problem due to more model parameters.
The results show that our model does not depend on a larger memory capacity.
In this paper, we adopt the dimension of $64$ by default.

\subsection{More adaptation results}
\vspace{-0.5mm}
\label{sup:supp_more_exp_results}
As mentioned in Sec.~\ref{sup:dataset_detail}, we perform adaptation experiments with a 60s adaptation set, and the quantitative results are shown in Tab.~\ref{tab:adaptation_60s} where our MemFace exhibits an obviously better performance in all aspects. 
Additionally, we provide several demo videos (5 minutes in total) to present our synthesized results and make comparisons with the state-of-the-art methods.

\subsection{Toy experiments}
\paragraph{Set the implicit memory to random values.}
To verify that the implicit memory complements some missing information, after training, we reset the keys and values in the implicit memory to values with normal distribution, and the results are shown in Fig.~\ref{fig:supp_toy-a2e}.  
We can see that complementing random information leads to a decrease in lip-sync quality.

\paragraph{Set the explicit memory to the visual appearance of another person.}
To verify that the explicit memory complements useful missing information, we set the explicit memory to the visual appearance of another person and the results are shown in Fig.~\ref{fig:supp_toy-render}. We can see that complementing another's pixel-level details leads to a decrease both in lip-sync and rendering quality.

\end{document}


\title{Memories are One-to-Many Mapping Alleviators in Talking Face Generation:\\
Supplementary Materials}

\author{First Author\\
Institution1\\
Institution1 address\\
{\tt\small firstauthor@i1.org}
\and
Second Author\\
Institution2\\
First line of institution2 address\\
{\tt\small secondauthor@i2.org}
}
\maketitle

\vspace{-1cm}
In this supplementary, we provide more details and results to make our paper more comprehensive. Specifically, we present more method details in Sec.~\ref{sec:method}, more experimental details in Sec.~\ref{sec:exp_detail}, and more experimental results in Sec.~\ref{sec:exp_results}.

\section{Method Details}
\label{sec:method}
\subsection{Construction of the explicit memory}
As illustrated in Sec. 3.3 of the main paper, to construct the explicit memory in neural-rendering model, we identify $N$ most dissimilar mouth shapes by calculating the Root-Mean-Square distance between mouth-related vertices, and then use the $N$ vertex-image pairs to build the explicit memory. 
Obviously, exploring all possible cases is almost infeasible due to heavy computation.
So we adopt a manually-designed algorithm to construct the explicit memory and the detailed steps are shown in Alg.~\ref{alg:expl_mem}.

\section{Experiment Details}
\label{sec:exp_detail}
\subsection{Dataset details}
\label{sec:dataset_detail}
\paragraph{Adaptation dataset.}
As described in Sec. 4.1 of the main paper, we randomly collect 10 speakers with various talking styles from HDTF dataset\cite{zhang2021flow}, and conduct two sets of experiments using adaptation data with different durations (15s and 30s). 
We additionally conduct a set of experiments with 60s adaptation set for cases with intense head movements, and the results are shown in Sec.~\ref{sec:supp_more_exp_results}.

\subsection{Training details}
We adopt Adam\cite{kingma2014adam} optimizer with a learning rate of $1$e$-4$ in the training stage. For adaptation experiments, we adapt the audio-to-expression model with a learning rate of $5$e$-6$ (200 epochs) and the neural-rendering model with a learning rate of $1$e$-4$ (50 epochs). 
To make the training easier, for the implicit memory, we update model parameters and memory alternately in the first half of training. 

\begin{algorithm}[H]
    \caption{Construction of the explicit memory.}
    \label{alg:expl_mem}
    \KwIn{A train set $T$ containing portrait images.}
    \KwOut{$N$ vertex-image pairs ($\rmK^{nr}$ and $\rmV^{nr}$).}
    
    \begin{algorithmic}[1]
    \STATE Set $N$ and randomly select $N$ vertex-image pairs in $T$ to
    initialize $\rmK^{nr}$ and $\rmV^{nr}$ of the explicit memory; 
    \STATE Define $D_{min}=f(\rmK^{nr})$, where $f$ is the function calculating Root-Mean-Square (RMS) distance between mouth-related vertices in $\rmK^{nr}$ to identify the two mouth shapes $\rmK_{m1}$ and $\rmK_{m2}$ that are most similar to each other, and $D_{min}$ is the corresponding RMS distance;
    \REPEAT
        \STATE $F$ = False;
        \FOR{($\rmK_{tmp}$, $\rmV_{tmp}$) in $T$}
            \STATE $\rmK_{m1}=\rmK_{tmp}$ to get $\rmK_{1}^{nr}$, $D_{m1}=f(\rmK_{1}^{nr})$;
            \STATE $\rmK_{m2}=\rmK_{tmp}$ to get $\rmK_{2}^{nr}$, $D_{m2}=f(\rmK_{2}^{nr})$;
            \IF{$\max(D_{m1}, D_{m2}) \textgreater D_{min}$}
            \STATE $m = \arg\max\limits_{m\in [m1,m2]}{D_{m}}$;
            \STATE $(\rmK_m,\rmV_m) = (\rmK_{tmp},\rmV_{tmp})$ to get $\rmK^{nr}$ and $\rmV^{nr}$;
            \STATE Calculate $f(\rmK^{nr})$ to update $\rmK_{m1}$,$\rmK_{m2}$ and $D_{min}$;
            \STATE $F$ = True;
            
            \ENDIF 
        \ENDFOR
    \UNTIL{$F$ = False}
    \STATE Return $\rmK^{nr}$ and $\rmV^{nr}$ as the $N$ vertex-image pairs to construct the explicit memory.
    \end{algorithmic}

\end{algorithm}

\begin{figure*}[t]
  \centering
  \includegraphics[width=1\linewidth]{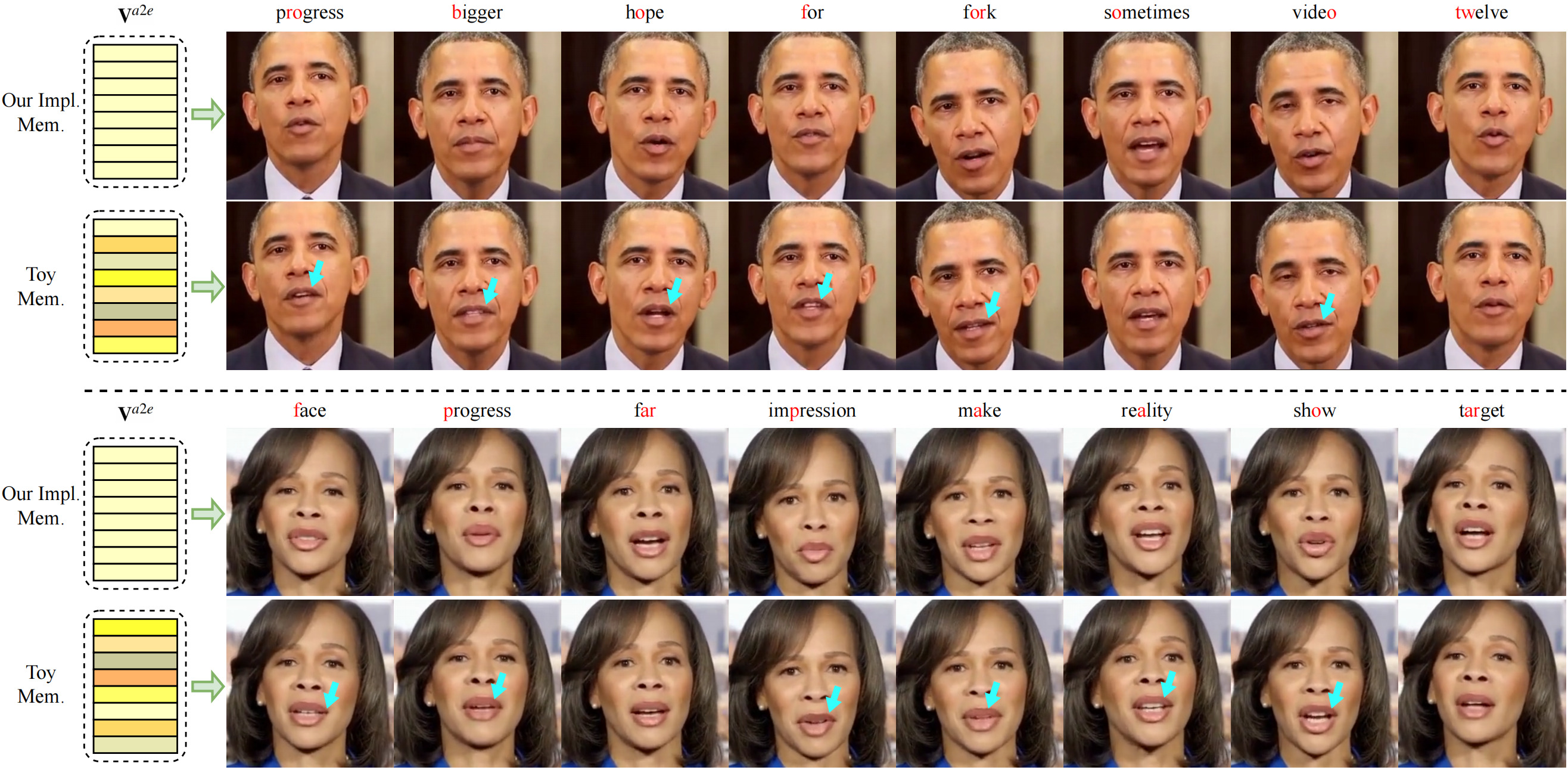}
  \caption{Set the implicit memory to random values. (upper half: train the audio-to-exppression model on GRID\cite{cooke2006audio}, lower half: adaption on HDTF\cite{zhang2021flow}). The blue arrows indicate obvious inferior lip-sync quality.}
  \vspace{-0.4cm}
  \label{fig:supp_toy-a2e}
\end{figure*}

\begin{figure*}[t]
  \centering
  \includegraphics[width=1\linewidth]{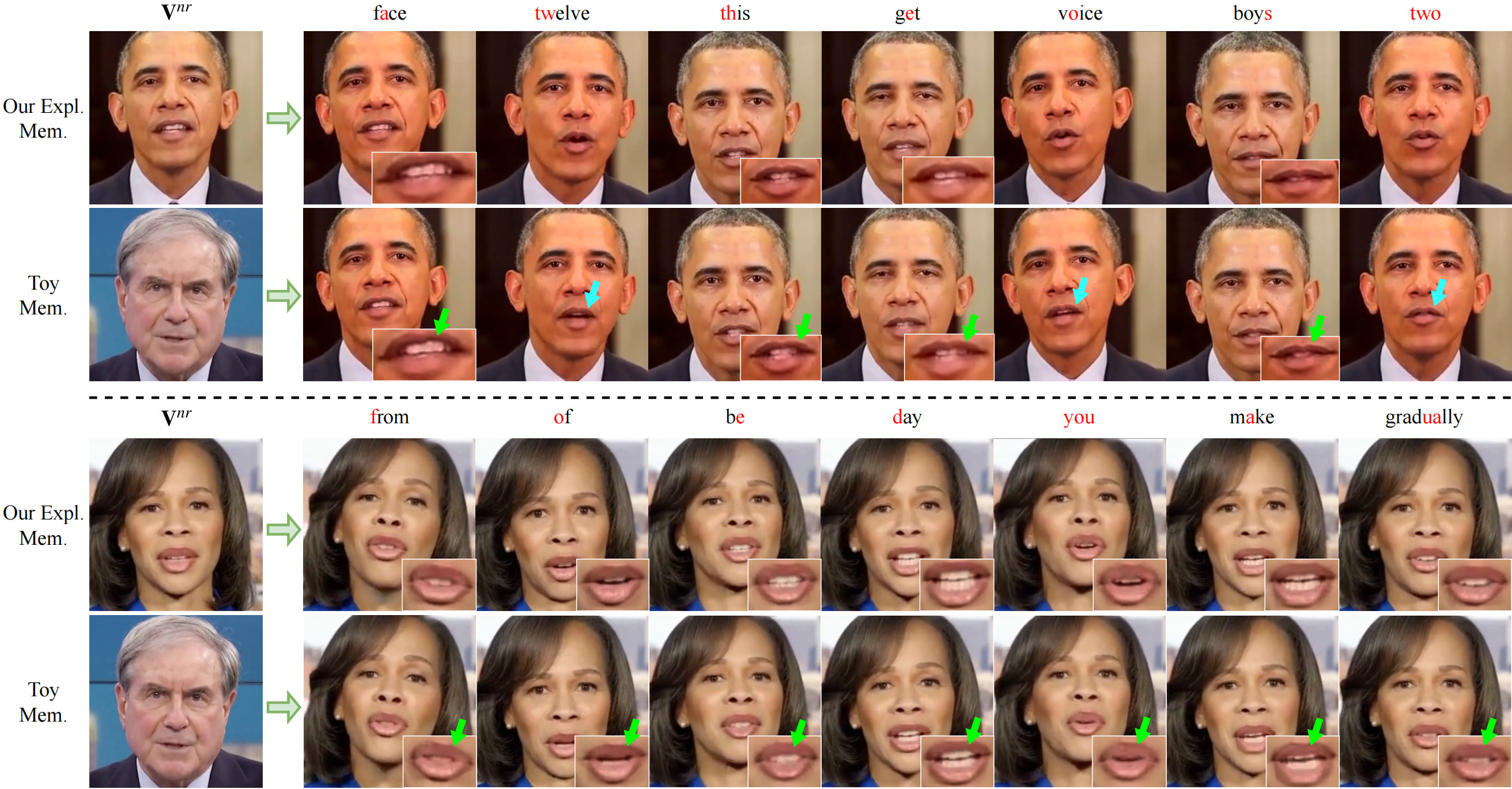}
  \caption{Set the explicit memory to visual appearance of another person. (upper half: train the neural-rendering model on Obama~\cite{suwajanakorn2017synthesizing}, lower half: adaptation on HDTF\cite{zhang2021flow}). The blue and green arrows indicate inferior lip-sync and rendering quality respectively.}
  \vspace{-0.5cm}
  \label{fig:supp_toy-render}
\end{figure*}

\subsection{Metrics details}
\paragraph{Subjective evaluation organization.}
For subjective evaluation, 20 experienced users are invited to participate in. 
We use MOS (Mean Opinion Score) and CMOS (Comparison Mean Opinion Score) as our metrics in comparison experiments and ablation studies respectively. 
In MOS experiments,we present one video at a time and ask the users to rate the presented video at five grades in lip-sync, rendering and overall quality respectively. 
Different grade choices in MOS are: Very Good (5), Good (4), Average (3), Poor (2), Very Poor (1). 
In CMOS experiments,
we present two videos at a time and ask the users to make
comparisons in lip-sync or rendering quality. 
Different grade choices in CMOS are: left video is obviously better (2), left video is a little better (1), indistinguishable (0), right video
is a little better (-1), right video is obviously better (-2).

We conduct online evaluation with well-designed Google questionnaires.
A detailed guideline with example videos is provided at the beginning of questionnaires to ensure a consistent grading criterion.

\begin{table*}[tb]
\setlength\tabcolsep{8pt}
\begin{center}
\small
\caption{More adaptation results.}
    \begin{tabular}{clccccc}
    \toprule[1.5pt]
    \multirow{2}{*}{Adaption Set} & \multirow{2}{*}{Method} &
    \multicolumn{3}{c}{Subjective Evaluation} & \multicolumn{2}{c}{Objective Evaluation} \\
    \cmidrule(l){3-5}
    \cmidrule(l){6-7}
    & & Lip-sync $\uparrow$ & Render$\uparrow$ & Overall $\uparrow$  & Sync-C $\uparrow$ & Sync-D $\downarrow$ \\
    \midrule
    \multirow{4}{*}{60s} & AD-NeRF\cite{guo2021ad}  & \underline{2.4347} & 2.2028 & 2.3477 & 1.677 & 13.049\\
    & DFRF\cite{shen2022learning} & 2.1738 & \underline{2.4637} & \underline{2.3767} & \underline{2.235} & \underline{12.170}\\
    & MemGAN\cite{yi2020audio}  & 1.6790 & 2.3187 & 1.7937 & 1.710& 12.872\\
    \midrule
    & Ours & \textbf{4.2746} & \textbf{3.9853}  & \textbf{4.2014} & \textbf{5.643} & \textbf{8.874} \\
    \bottomrule[1.5pt]
    \end{tabular}
    \vspace{-6mm}
    \label{tab:adaptation_60s}
\end{center}
\end{table*}

\section{More Experimental Results}
\label{sec:exp_results}
\subsection{More ablation studies}
\label{sec:supp_more_abla_studies}

\paragraph{Dimension of the implicit memory.}
In Sec. 4.3 of the main paper, we conduct ablation studies on the number of key-value pairs in the implicit memory. As a supplement, we also explore the dimension selection of keys and values in the implicit memory. In Tab.~\ref{tab:abla_size}, we set the dimension of keys and values to 32, 64 (our adoption), 96 and 128 respectively and show the corresponding performance. From Tab.~\ref{tab:abla_size}, with the increase of dimensions, the prediction performance on GRID\cite{cooke2006audio} dataset (RMSE-1) increases while the performance on HDTF\cite{zhang2021flow} adaptation set (RMSE-2) decreases which may be caused by an overfitting problem due to more model parameters.
The results show that our model does not depend on larger memory capacity.
In this paper, we adopt the dimension of 64 with an overall consideration.


\begin{table}[t]
\caption{Ablation study on the size of the implicit memory. The number of key-value pairs is 1000, and $D$ means the dimension of keys and values.
Param. indicates the number of parameters of the audio-to-expression model.
We perform training with GRID\cite{cooke2006audio} data (RMSE-1) and perform adaptation with HDTF\cite{zhang2021flow} data (RMSE-2) to study the effect of $D$.
The bold font is our adoption.
}
\vspace{-1mm}
\small
\centering
\label{tab:abla_size}
    \begin{tabular}[t]{@{}l|ccc@{}}
	    \toprule[1.5pt]
		$D$  & Param. & RMSE-1$\downarrow$ &  RMSE-2$\downarrow$  \\
		\midrule
		32  & 178k  & 0.0832  &  0.1341  \\
		\textbf{64}  &  \textbf{240k}   & \textbf{ 0.0826} &  \textbf{0.1343}  \\
		96  &  314k   &   0.0824  &  0.1352   \\
		128  &  382k  &  0.0823  &  0.1357  \\
		\bottomrule[1.5pt]
	\end{tabular}
\end{table}


\subsection{More experimental results}
\vspace{-0.5mm}
\label{sec:supp_more_exp_results}
As mentioned in Sec.~\ref{sec:dataset_detail}, we perform adaptation experiments with 60s adaptation set, and the quantitative results are shown in Tab.~\ref{tab:adaptation_60s} where our method exhibits an obviously better performance in all aspects. 
Additionally, we provide several demo videos (5 minutes in total) to present our synthesized results and make comparison with the state-of-the-art methods.

\subsection{Toy experiments}
\paragraph{Set the implicit memory to random values.}
To verify that the implicit memory complements some missing information, after training, we reset the keys and values in the implicit memory to values with normal distribution, and the results are shown in Fig.~\ref{fig:supp_toy-a2e}. 
We can see that complementing random information leads to a decrease in lip-sync quality.

\paragraph{Set the explicit memory to visual appearance of another person.}
To verify that the explicit memory complements useful missing information, we set the explicit memory to visual appearance of another person and the results are shown in Fig.~\ref{fig:supp_toy-render}. We can see that complementing another's pixel-level details leads to a decrease both in lip-sync and rendering quality.


{\small
\bibliographystyle{ieee_fullname}
\bibliography{egbib}
}

%% file: math_commands.tex
\usepackage{amsmath,amsfonts,bm}

\def\eqref#1{equation~\ref{#1}}

\def\1{\bm{1}}

\def\rmA{{\mathbf{A}}}

\def\rmF{{\mathbf{F}}}

\def\rmI{{\mathbf{I}}}

\def\rmK{{\mathbf{K}}}

\def\rmO{{\mathbf{O}}}

\def\rmQ{{\mathbf{Q}}}

\def\rmV{{\mathbf{V}}}
\def\rmW{{\mathbf{W}}}
\def\rmX{{\mathbf{X}}}

\DeclareMathAlphabet{\mathsfit}{\encodingdefault}{\sfdefault}{m}{sl}
\SetMathAlphabet{\mathsfit}{bold}{\encodingdefault}{\sfdefault}{bx}{n}

%% file: arxiv.bbl
\begin{thebibliography}{10}\itemsep=-1pt

\bibitem{agarwal2020detecting}
Shruti Agarwal, Hany Farid, Ohad Fried, and Maneesh Agrawala.
\newblock Detecting deep-fake videos from phoneme-viseme mismatches.
\newblock In {\em Proceedings of the IEEE/CVF conference on computer vision and
  pattern recognition workshops}, pages 660--661, 2020.

\bibitem{amodei2016deep}
Dario Amodei, Sundaram Ananthanarayanan, Rishita Anubhai, Jingliang Bai, Eric
  Battenberg, Carl Case, Jared Casper, Bryan Catanzaro, Qiang Cheng, Guoliang
  Chen, et~al.
\newblock Deep speech 2: End-to-end speech recognition in english and mandarin.
\newblock In {\em International conference on machine learning}, pages
  173--182. PMLR, 2016.

\bibitem{blanz1999morphable}
Volker Blanz and Thomas Vetter.
\newblock A morphable model for the synthesis of 3d faces.
\newblock In {\em Proceedings of the 26th annual conference on Computer
  graphics and interactive techniques}, pages 187--194, 1999.

\bibitem{blattmann2022retrieval}
Andreas Blattmann, Robin Rombach, Kaan Oktay, and Bj{\"o}rn Ommer.
\newblock Retrieval-augmented diffusion models.
\newblock {\em arXiv preprint arXiv:2204.11824}, 2022.

\bibitem{casanova2022yourtts}
Edresson Casanova, Julian Weber, Christopher~D Shulby, Arnaldo~Candido Junior,
  Eren G{\"o}lge, and Moacir~A Ponti.
\newblock Yourtts: Towards zero-shot multi-speaker tts and zero-shot voice
  conversion for everyone.
\newblock In {\em International Conference on Machine Learning}, pages
  2709--2720. PMLR, 2022.

\bibitem{chen2020talking}
Lele Chen, Guofeng Cui, Celong Liu, Zhong Li, Ziyi Kou, Yi Xu, and Chenliang
  Xu.
\newblock Talking-head generation with rhythmic head motion.
\newblock In {\em European Conference on Computer Vision}, pages 35--51.
  Springer, 2020.

\bibitem{chen2019hierarchical}
Lele Chen, Ross~K Maddox, Zhiyao Duan, and Chenliang Xu.
\newblock Hierarchical cross-modal talking face generation with dynamic
  pixel-wise loss.
\newblock In {\em Proceedings of the IEEE/CVF conference on computer vision and
  pattern recognition}, pages 7832--7841, 2019.

\bibitem{chen2022transformer}
Liyang Chen, Zhiyong Wu, Jun Ling, Runnan Li, Xu Tan, and Sheng Zhao.
\newblock Transformer-s2a: Robust and efficient speech-to-animation.
\newblock In {\em ICASSP 2022-2022 IEEE International Conference on Acoustics,
  Speech and Signal Processing (ICASSP)}, pages 7247--7251. IEEE, 2022.

\bibitem{chen2020duallip}
Weicong Chen, Xu Tan, Yingce Xia, Tao Qin, Yu Wang, and Tie-Yan Liu.
\newblock Duallip: A system for joint lip reading and generation.
\newblock In {\em Proceedings of the 28th ACM International Conference on
  Multimedia}, pages 1985--1993, 2020.

\bibitem{chung2016out}
Joon~Son Chung and Andrew Zisserman.
\newblock Out of time: automated lip sync in the wild.
\newblock In {\em Asian conference on computer vision}, pages 251--263.
  Springer, 2016.

\bibitem{cooke2006audio}
Martin Cooke, Jon Barker, Stuart Cunningham, and Xu Shao.
\newblock An audio-visual corpus for speech perception and automatic speech
  recognition.
\newblock {\em The Journal of the Acoustical Society of America},
  120(5):2421--2424, 2006.

\bibitem{cudeiro2019capture}
Daniel Cudeiro, Timo Bolkart, Cassidy Laidlaw, Anurag Ranjan, and Michael~J
  Black.
\newblock Capture, learning, and synthesis of 3d speaking styles.
\newblock In {\em Proceedings of the IEEE/CVF Conference on Computer Vision and
  Pattern Recognition}, pages 10101--10111, 2019.

\bibitem{deng2019arcface}
Jiankang Deng, Jia Guo, Niannan Xue, and Stefanos Zafeiriou.
\newblock Arcface: Additive angular margin loss for deep face recognition.
\newblock In {\em Proceedings of the IEEE/CVF conference on computer vision and
  pattern recognition}, pages 4690--4699, 2019.

\bibitem{dinh2016density}
Laurent Dinh, Jascha Sohl-Dickstein, and Samy Bengio.
\newblock Density estimation using real nvp.
\newblock In {\em International Conference on Learning Representations}, 2017.

\bibitem{gafni2021dynamic}
Guy Gafni, Justus Thies, Michael Zollhofer, and Matthias Nie{\ss}ner.
\newblock Dynamic neural radiance fields for monocular 4d facial avatar
  reconstruction.
\newblock In {\em Proceedings of the IEEE/CVF Conference on Computer Vision and
  Pattern Recognition}, pages 8649--8658, 2021.

\bibitem{goodfellow2014generative}
Ian Goodfellow, Jean Pouget-Abadie, Mehdi Mirza, Bing Xu, David Warde-Farley,
  Sherjil Ozair, Aaron Courville, and Yoshua Bengio.
\newblock Generative adversarial nets.
\newblock In {\em Advances in Neural Information Processing Systems (NIPS)},
  pages 2672--2680, 2014.

\bibitem{guo2021ad}
Yudong Guo, Keyu Chen, Sen Liang, Yong-Jin Liu, Hujun Bao, and Juyong Zhang.
\newblock Ad-nerf: Audio driven neural radiance fields for talking head
  synthesis.
\newblock In {\em Proceedings of the IEEE/CVF International Conference on
  Computer Vision}, pages 5784--5794, 2021.

\bibitem{han2022show}
Ligong Han, Jian Ren, Hsin-Ying Lee, Francesco Barbieri, Kyle Olszewski,
  Shervin Minaee, Dimitris Metaxas, and Sergey Tulyakov.
\newblock Show me what and tell me how: Video synthesis via multimodal
  conditioning.
\newblock In {\em Proceedings of the IEEE/CVF Conference on Computer Vision and
  Pattern Recognition}, pages 3615--3625, 2022.

\bibitem{he2016deep}
Kaiming He, Xiangyu Zhang, Shaoqing Ren, and Jian Sun.
\newblock Deep residual learning for image recognition.
\newblock In {\em Proceedings of the IEEE conference on computer vision and
  pattern recognition}, pages 770--778, 2016.

\bibitem{hong2022depth}
Fa-Ting Hong, Longhao Zhang, Li Shen, and Dan Xu.
\newblock Depth-aware generative adversarial network for talking head video
  generation.
\newblock In {\em Proceedings of the IEEE/CVF Conference on Computer Vision and
  Pattern Recognition}, pages 3397--3406, 2022.

\bibitem{ji2021audio}
Xinya Ji, Hang Zhou, Kaisiyuan Wang, Wayne Wu, Chen~Change Loy, Xun Cao, and
  Feng Xu.
\newblock Audio-driven emotional video portraits.
\newblock In {\em Proceedings of the IEEE/CVF conference on computer vision and
  pattern recognition}, pages 14080--14089, 2021.

\bibitem{johnson2016perceptual}
Justin Johnson, Alexandre Alahi, and Li Fei-Fei.
\newblock Perceptual losses for real-time style transfer and super-resolution.
\newblock In {\em European conference on computer vision}, pages 694--711.
  Springer, 2016.

\bibitem{kazemi2018unsupervised}
Hadi Kazemi, Sobhan Soleymani, Fariborz Taherkhani, Seyed Iranmanesh, and
  Nasser Nasrabadi.
\newblock Unsupervised image-to-image translation using domain-specific
  variational information bound.
\newblock {\em Advances in neural information processing systems}, 31, 2018.

\bibitem{khandelwal2021nearest}
Urvashi Khandelwal, Angela Fan, Dan Jurafsky, Luke Zettlemoyer, and Mike Lewis.
\newblock Nearest neighbor machine translation.
\newblock In {\em International Conference on Learning Representations}, 2021.

\bibitem{khandelwal2020generalization}
Urvashi Khandelwal, Omer Levy, Dan Jurafsky, Luke Zettlemoyer, and Mike Lewis.
\newblock Generalization through memorization: Nearest neighbor language
  models.
\newblock In {\em International Conference on Learning Representations}, 2020.

\bibitem{kim2018deep}
Hyeongwoo Kim, Pablo Garrido, Ayush Tewari, Weipeng Xu, Justus Thies, Matthias
  Niessner, Patrick P{\'e}rez, Christian Richardt, Michael Zollh{\"o}fer, and
  Christian Theobalt.
\newblock Deep video portraits.
\newblock {\em ACM Transactions on Graphics (TOG)}, 37(4):1--14, 2018.

\bibitem{kingma2014adam}
Diederik~P Kingma and Jimmy Ba.
\newblock Adam: A method for stochastic optimization.
\newblock In {\em Proceedings of the International Conference on Learning
  Representations}, 2015.

\bibitem{kingma2013auto}
Diederik~P Kingma and Max Welling.
\newblock Auto-encoding variational bayes.
\newblock In {\em International Conference on Learning Representations}, 2014.

\bibitem{kumar2016ask}
Ankit Kumar, Ozan Irsoy, Peter Ondruska, Mohit Iyyer, James Bradbury, Ishaan
  Gulrajani, Victor Zhong, Romain Paulus, and Richard Socher.
\newblock Ask me anything: Dynamic memory networks for natural language
  processing.
\newblock In {\em International conference on machine learning}, pages
  1378--1387. PMLR, 2016.

\bibitem{lahiri2021lipsync3d}
Avisek Lahiri, Vivek Kwatra, Christian Frueh, John Lewis, and Chris Bregler.
\newblock Lipsync3d: Data-efficient learning of personalized 3d talking faces
  from video using pose and lighting normalization.
\newblock In {\em Proceedings of the IEEE/CVF conference on computer vision and
  pattern recognition}, pages 2755--2764, 2021.

\bibitem{lee2018memory}
Sangho Lee, Jinyoung Sung, Youngjae Yu, and Gunhee Kim.
\newblock A memory network approach for story-based temporal summarization of
  360 videos.
\newblock In {\em Proceedings of the IEEE Conference on Computer Vision and
  Pattern Recognition}, pages 1410--1419, 2018.

\bibitem{li2021audio2gestures}
Jing Li, Di Kang, Wenjie Pei, Xuefei Zhe, Ying Zhang, Zhenyu He, and Linchao
  Bao.
\newblock Audio2gestures: Generating diverse gestures from speech audio with
  conditional variational autoencoders.
\newblock In {\em Proceedings of the IEEE/CVF International Conference on
  Computer Vision}, pages 11293--11302, 2021.

\bibitem{liang2022expressive}
Borong Liang, Yan Pan, Zhizhi Guo, Hang Zhou, Zhibin Hong, Xiaoguang Han, Junyu
  Han, Jingtuo Liu, Errui Ding, and Jingdong Wang.
\newblock Expressive talking head generation with granular audio-visual
  control.
\newblock In {\em Proceedings of the IEEE/CVF Conference on Computer Vision and
  Pattern Recognition}, pages 3387--3396, 2022.

\bibitem{ling2022stableface}
Jun Ling, Xu Tan, Liyang Chen, Runnan Li, Yuchao Zhang, Sheng Zhao, and Li
  Song.
\newblock Stableface: Analyzing and improving motion stability for talking face
  generation.
\newblock {\em arXiv preprint arXiv:2208.13717}, 2022.

\bibitem{liu2022semantic}
Xian Liu, Yinghao Xu, Qianyi Wu, Hang Zhou, Wayne Wu, and Bolei Zhou.
\newblock Semantic-aware implicit neural audio-driven video portrait
  generation.
\newblock In {\em European Conference on Computer Vision}. Springer, 2022.

\bibitem{long2022retrieval}
Alexander Long, Wei Yin, Thalaiyasingam Ajanthan, Vu Nguyen, Pulak Purkait,
  Ravi Garg, Alan Blair, Chunhua Shen, and Anton van~den Hengel.
\newblock Retrieval augmented classification for long-tail visual recognition.
\newblock In {\em Proceedings of the IEEE/CVF Conference on Computer Vision and
  Pattern Recognition}, pages 6959--6969, 2022.

\bibitem{lu2021live}
Yuanxun Lu, Jinxiang Chai, and Xun Cao.
\newblock Live speech portraits: real-time photorealistic talking-head
  animation.
\newblock {\em ACM Transactions on Graphics (TOG)}, 40(6):1--17, 2021.

\bibitem{meshry2021learned}
Moustafa Meshry, Saksham Suri, Larry~S Davis, and Abhinav Shrivastava.
\newblock Learned spatial representations for few-shot talking-head synthesis.
\newblock In {\em Proceedings of the IEEE/CVF International Conference on
  Computer Vision}, pages 13829--13838, 2021.

\bibitem{mildenhall2021nerf}
Ben Mildenhall, Pratul~P Srinivasan, Matthew Tancik, Jonathan~T Barron, Ravi
  Ramamoorthi, and Ren Ng.
\newblock Nerf: Representing scenes as neural radiance fields for view
  synthesis.
\newblock {\em Communications of the ACM}, 65(1):99--106, 2021.

\bibitem{miller2016key}
Alexander Miller, Adam Fisch, Jesse Dodge, Amir-Hossein Karimi, Antoine Bordes,
  and Jason Weston.
\newblock Key-value memory networks for directly reading documents.
\newblock In {\em Proceedings of the 2016 Conference on Empirical Methods in
  Natural Language Processing}, pages 1400--1409, 2016.

\bibitem{park2022synctalkface}
Se~Jin Park, Minsu Kim, Joanna Hong, Jeongsoo Choi, and Yong~Man Ro.
\newblock Synctalkface: Talking face generation with precise lip-syncing via
  audio-lip memory.
\newblock In {\em 36th AAAI Conference on Artificial Intelligence (AAAI 22)}.
  Association for the Advancement of Artificial Intelligence, 2022.

\bibitem{prajwal2020lip}
KR Prajwal, Rudrabha Mukhopadhyay, Vinay~P Namboodiri, and CV Jawahar.
\newblock A lip sync expert is all you need for speech to lip generation in the
  wild.
\newblock In {\em Proceedings of the 28th ACM International Conference on
  Multimedia}, pages 484--492, 2020.

\bibitem{qian2021speech}
Shenhan Qian, Zhi Tu, Yihao Zhi, Wen Liu, and Shenghua Gao.
\newblock Speech drives templates: Co-speech gesture synthesis with learned
  templates.
\newblock In {\em Proceedings of the IEEE/CVF International Conference on
  Computer Vision}, pages 11077--11086, 2021.

\bibitem{ronneberger2015u}
Olaf Ronneberger, Philipp Fischer, and Thomas Brox.
\newblock U-net: Convolutional networks for biomedical image segmentation.
\newblock In {\em International Conference on Medical image computing and
  computer-assisted intervention}, pages 234--241. Springer, 2015.

\bibitem{shen2022learning}
Shuai Shen, Wanhua Li, Zheng Zhu, Yueqi Duan, Jie Zhou, and Jiwen Lu.
\newblock Learning dynamic facial radiance fields for few-shot talking head
  synthesis.
\newblock In {\em European Conference on Computer Vision}. Springer, 2022.

\bibitem{shen2020one}
Zengming Shen, S~Kevin Zhou, Yifan Chen, Bogdan Georgescu, Xuqi Liu, and Thomas
  Huang.
\newblock One-to-one mapping for unpaired image-to-image translation.
\newblock In {\em Proceedings of the IEEE/CVF Winter Conference on Applications
  of Computer Vision}, pages 1170--1179, 2020.

\bibitem{sheynin2022knn}
Shelly Sheynin, Oron Ashual, Adam Polyak, Uriel Singer, Oran Gafni, Eliya
  Nachmani, and Yaniv Taigman.
\newblock Knn-diffusion: Image generation via large-scale retrieval.
\newblock {\em arXiv preprint arXiv:2204.02849}, 2022.

\bibitem{siarohin2019first}
Aliaksandr Siarohin, St{\'e}phane Lathuili{\`e}re, Sergey Tulyakov, Elisa
  Ricci, and Nicu Sebe.
\newblock First order motion model for image animation.
\newblock {\em Advances in Neural Information Processing Systems}, 32, 2019.

\bibitem{siddiqui2021retrievalfuse}
Yawar Siddiqui, Justus Thies, Fangchang Ma, Qi Shan, Matthias Nie{\ss}ner, and
  Angela Dai.
\newblock Retrievalfuse: Neural 3d scene reconstruction with a database.
\newblock In {\em Proceedings of the IEEE/CVF International Conference on
  Computer Vision}, pages 12568--12577, 2021.

\bibitem{smith2020morphable}
William~AP Smith, Alassane Seck, Hannah Dee, Bernard Tiddeman, Joshua~B
  Tenenbaum, and Bernhard Egger.
\newblock A morphable face albedo model.
\newblock In {\em Proceedings of the IEEE/CVF Conference on Computer Vision and
  Pattern Recognition}, pages 5011--5020, 2020.

\bibitem{sukhbaatar2015end}
Sainbayar Sukhbaatar, Jason Weston, Rob Fergus, et~al.
\newblock End-to-end memory networks.
\newblock {\em Advances in neural information processing systems}, 28, 2015.

\bibitem{sun2016phonetic}
Lifa Sun, Kun Li, Hao Wang, Shiyin Kang, and Helen Meng.
\newblock Phonetic posteriorgrams for many-to-one voice conversion without
  parallel data training.
\newblock In {\em 2016 IEEE International Conference on Multimedia and Expo
  (ICME)}, pages 1--6. IEEE, 2016.

\bibitem{suwajanakorn2017synthesizing}
Supasorn Suwajanakorn, Steven~M Seitz, and Ira Kemelmacher-Shlizerman.
\newblock Synthesizing obama: learning lip sync from audio.
\newblock {\em ACM Transactions on Graphics (ToG)}, 36(4):1--13, 2017.

\bibitem{tan2022naturalspeech}
Xu Tan, Jiawei Chen, Haohe Liu, Jian Cong, Chen Zhang, Yanqing Liu, Xi Wang,
  Yichong Leng, Yuanhao Yi, Lei He, et~al.
\newblock Naturalspeech: End-to-end text to speech synthesis with human-level
  quality.
\newblock {\em arXiv preprint arXiv:2205.04421}, 2022.

\bibitem{tang2022real}
Jiaxiang Tang, Kaisiyuan Wang, Hang Zhou, Xiaokang Chen, Dongliang He, Tianshu
  Hu, Jingtuo Liu, Gang Zeng, and Jingdong Wang.
\newblock Real-time neural radiance talking portrait synthesis via
  audio-spatial decomposition.
\newblock {\em arXiv preprint arXiv:2211.12368}, 2022.

\bibitem{thies2020neural}
Justus Thies, Mohamed Elgharib, Ayush Tewari, Christian Theobalt, and Matthias
  Nie{\ss}ner.
\newblock Neural voice puppetry: Audio-driven facial reenactment.
\newblock In {\em European conference on computer vision}, pages 716--731.
  Springer, 2020.

\bibitem{tseng2020retrievegan}
Hung-Yu Tseng, Hsin-Ying Lee, Lu Jiang, Ming-Hsuan Yang, and Weilong Yang.
\newblock Retrievegan: Image synthesis via differentiable patch retrieval.
\newblock In {\em European Conference on Computer Vision}, pages 242--257.
  Springer, 2020.

\bibitem{vaswani2017attention}
Ashish Vaswani, Noam Shazeer, Niki Parmar, Jakob Uszkoreit, Llion Jones,
  Aidan~N Gomez, {\L}ukasz Kaiser, and Illia Polosukhin.
\newblock Attention is all you need.
\newblock {\em Advances in neural information processing systems}, 30, 2017.

\bibitem{wang2021audio2head}
Suzhen Wang, Lincheng Li, Yu Ding, Changjie Fan, and Xin Yu.
\newblock Audio2head: Audio-driven one-shot talking-head generation with
  natural head motion.
\newblock {\em arXiv preprint arXiv:2107.09293}, 2021.

\bibitem{wang2022low}
Yufei Wang, Renjie Wan, Wenhan Yang, Haoliang Li, Lap-Pui Chau, and Alex Kot.
\newblock Low-light image enhancement with normalizing flow.
\newblock In {\em Proceedings of the AAAI Conference on Artificial
  Intelligence}, volume~36, pages 2604--2612, 2022.

\bibitem{wang2018three}
Yining Wang, Jiajun Zhang, Feifei Zhai, Jingfang Xu, and Chengqing Zong.
\newblock Three strategies to improve one-to-many multilingual translation.
\newblock In {\em Proceedings of the 2018 Conference on Empirical Methods in
  Natural Language Processing}, pages 2955--2960, 2018.

\bibitem{wen2020photorealistic}
Xin Wen, Miao Wang, Christian Richardt, Ze-Yin Chen, and Shi-Min Hu.
\newblock Photorealistic audio-driven video portraits.
\newblock {\em IEEE Transactions on Visualization and Computer Graphics},
  26(12):3457--3466, 2020.

\bibitem{weston2014memory}
Jason Weston, Sumit Chopra, and Antoine Bordes.
\newblock Memory networks.
\newblock In {\em International Conference on Learning Representations}, 2015.

\bibitem{wood2021fake}
Erroll Wood, Tadas Baltru{\v{s}}aitis, Charlie Hewitt, Sebastian Dziadzio,
  Thomas~J Cashman, and Jamie Shotton.
\newblock Fake it till you make it: face analysis in the wild using synthetic
  data alone.
\newblock In {\em Proceedings of the IEEE/CVF international conference on
  computer vision}, pages 3681--3691, 2021.

\bibitem{wu2021imitating}
Haozhe Wu, Jia Jia, Haoyu Wang, Yishun Dou, Chao Duan, and Qingshan Deng.
\newblock Imitating arbitrary talking style for realistic audio-driven talking
  face synthesis.
\newblock In {\em Proceedings of the 29th ACM International Conference on
  Multimedia}, pages 1478--1486, 2021.

\bibitem{wu2022adaspeech}
Yihan Wu, Xu Tan, Bohan Li, Lei He, Sheng Zhao, Ruihua Song, Tao Qin, and
  Tie-Yan Liu.
\newblock Adaspeech 4: Adaptive text to speech in zero-shot scenarios.
\newblock {\em arXiv preprint arXiv:2204.00436}, 2022.

\bibitem{xin2021cross}
Detai Xin, Yuki Saito, Shinnosuke Takamichi, Tomoki Koriyama, and Hiroshi
  Saruwatari.
\newblock Cross-lingual speaker adaptation using domain adaptation and speaker
  consistency loss for text-to-speech synthesis.
\newblock In {\em Interspeech}, pages 1614--1618, 2021.

\bibitem{yi2020audio}
Ran Yi, Zipeng Ye, Juyong Zhang, Hujun Bao, and Yong-Jin Liu.
\newblock Audio-driven talking face video generation with learning-based
  personalized head pose.
\newblock {\em arXiv preprint arXiv:2002.10137}, 2020.

\bibitem{zhang2021facial}
Chenxu Zhang, Yifan Zhao, Yifei Huang, Ming Zeng, Saifeng Ni, Madhukar
  Budagavi, and Xiaohu Guo.
\newblock Facial: Synthesizing dynamic talking face with implicit attribute
  learning.
\newblock In {\em Proceedings of the IEEE/CVF international conference on
  computer vision}, pages 3867--3876, 2021.

\bibitem{zhang2018unreasonable}
Richard Zhang, Phillip Isola, Alexei~A Efros, Eli Shechtman, and Oliver Wang.
\newblock The unreasonable effectiveness of deep features as a perceptual
  metric.
\newblock In {\em Proceedings of the IEEE conference on computer vision and
  pattern recognition}, pages 586--595, 2018.

\bibitem{zhang2021flow}
Zhimeng Zhang, Lincheng Li, Yu Ding, and Changjie Fan.
\newblock Flow-guided one-shot talking face generation with a high-resolution
  audio-visual dataset.
\newblock In {\em Proceedings of the IEEE/CVF Conference on Computer Vision and
  Pattern Recognition}, pages 3661--3670, 2021.

\bibitem{zhou2021pose}
Hang Zhou, Yasheng Sun, Wayne Wu, Chen~Change Loy, Xiaogang Wang, and Ziwei
  Liu.
\newblock Pose-controllable talking face generation by implicitly modularized
  audio-visual representation.
\newblock In {\em Proceedings of the IEEE/CVF conference on computer vision and
  pattern recognition}, pages 4176--4186, 2021.

\end{thebibliography}
